\documentclass{article} % For LaTeX2e
\usepackage{iclr2025_conference,times}

\usepackage{amsmath,amsfonts,bm}

\def\eqref#1{equation~\ref{#1}}
\def\1{\bm{1}}

\DeclareMathAlphabet{\mathsfit}{\encodingdefault}{\sfdefault}{m}{sl}
\SetMathAlphabet{\mathsfit}{bold}{\encodingdefault}{\sfdefault}{bx}{n}

\def\gD{{\mathcal{D}}}

\def\gJ{{\mathcal{J}}}

\def\sX{{\mathbb{X}}}

\newcommand{\E}{\mathbb{E}}

\usepackage{hyperref}
\usepackage{url}

\usepackage{lipsum} 
\usepackage{graphicx}

\usepackage{subcaption}
\usepackage{caption}
\usepackage{wrapfig}

\usepackage{multirow}
\usepackage{adjustbox}

\usepackage{amsmath}
\usepackage{amsthm}
\usepackage{xcolor}
\newcommand{\csfo}{$\text{SF1}^c$}
\newcommand{\isfo}{$\text{SF1}^i$}
\newcommand{\nspace}{ }

\usepackage{algorithm}
\usepackage{algorithmic}
\usepackage{amsmath}

\usepackage{marginnote}
\usepackage{todonotes}
\usepackage{soul}

\title{Improving Unsupervised Constituency Parsing via Maximizing Semantic Information}

\author{Junjie Chen\textsuperscript{1}, Xiangheng He\textsuperscript{2}, Yusuke Miyao\textsuperscript{1}, Danushka Bollegala \textsuperscript{3} \\
Department of Computer Science, the University of Tokyo\textsuperscript{1}\\
GLAM – Group on Language, Audio, \& Music, Imperial College London\textsuperscript{2}\\
Department of Computer Science, the University of Liverpool\textsuperscript{3}\\
\texttt{christopher@is.s.u-tokyo.ac.jp, x.he20@imperial.ac.uk}\\
\texttt{yusuke@is.s.u-tokyo.ac.jp, danushka@liverpool.ac.uk} \\
}

\iclrfinalcopy

\begin{document}

\maketitle

\begin{abstract}
Unsupervised constituency parsers organize phrases within a sentence into a tree-shaped syntactic constituent structure that reflects the organization of sentence semantics. 
However, the traditional objective of maximizing sentence log-likelihood (LL) does not explicitly account for the close relationship between the constituent structure and the semantics, resulting in a weak correlation between LL values and parsing accuracy.
In this paper, we introduce a novel objective that trains parsers by maximizing SemInfo, the semantic information encoded in constituent structures.
We introduce a bag-of-substrings model to represent the semantics and estimate the SemInfo value using the probability-weighted information metric.
We apply the SemInfo maximization objective to training Probabilistic Context-Free Grammar (PCFG) parsers and develop a Tree Conditional Random Field (TreeCRF)-based model to facilitate the training. 
Experiments show that SemInfo correlates more strongly with parsing accuracy than LL, establishing SemInfo as a better unsupervised parsing objective.
As a result, our algorithm significantly improves parsing accuracy by an average of 7.85 sentence-F1 scores across five PCFG variants and in four languages, achieving state-of-the-art level results in three of the four languages.
\end{abstract}
% We propose a novel framework for estimating SemInfo by representing semantics with a bag-of-substring model and applying a probability-weighted information metric.

\section{Introduction}
\begin{figure}[t]
    \centering
    \includegraphics[width=\textwidth]{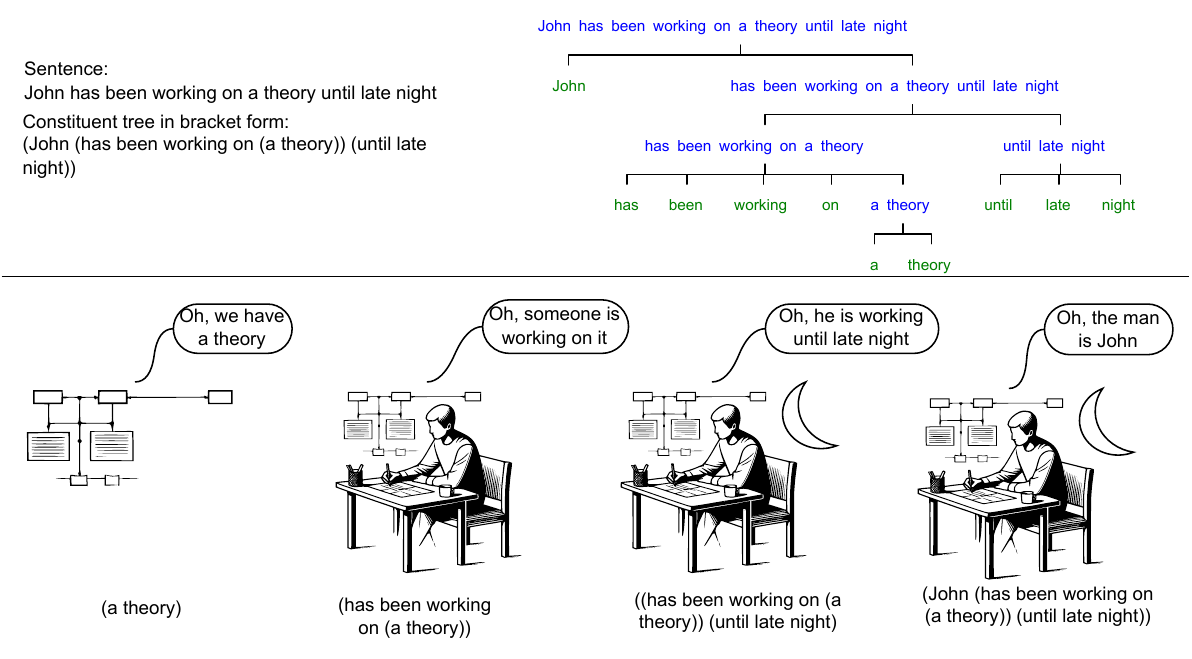}
    \caption{An illustration of the progressive semantics build-up in accordance with the constituent structure. The tree structure in the top-right shows the simplified constituent structure for illustration purposes. Constituent substrings are highlighted in blue.}
    \label{fig:substring-semantics}
\end{figure}

Unsupervised constituency parsing is a syntactic task of organizing phrases of a sentence into a tree-shaped constituent structure without relying on linguistic annotations \citep{klein-manning-2002-generative}.
The constituent structure is a fundamental tool in analyzing sentence semantics (i.e., the meaning) \citep{Carnie_2007, Steedman_2000}.
It can significantly improve performance for downstream Natural Language Processing systems, such as natural language inference \citep{he-etal-2020-enhancing}, machine translation \citep{xie-xing-2017-constituent} and semantic role labeling \citep{chen-etal-2022-modeling} systems.
It guides the progressive construction of the sentence semantics, as illustrated in Figure~\ref{fig:substring-semantics}. 
Each constituent in the structure corresponds to a meaningful substring, forming partial representations of the sentence semantics. 
One can easily recover the full sentence semantics by gradually constructing the semantic representation of those constituent substrings.
Following the observation, we hypothesize that \emph{constituent substrings in the sentence carry significant semantic information}.

Maximizing sentence log-likelihood has traditionally been the primary training objective for training unsupervised constituency parsers \citep{eisner-2016-inside, kim-etal-2019-compound}.
However, the Log-Likelihood (LL) function does not explicitly factor in the syntax-semantics alignment.
This leads to a poor correlation between the LL value and the parsing accuracy.
We will further discuss this poor correlation in Section~\ref{sec:seminfo-corr}.
As pointed out in previous research, it is challenging to train a Probabilistic Context-Free Grammar (PCFG) parser that outperforms trivial baselines with the LL maximization objective \citep{Carroll1992TwoEO, kim-etal-2019-compound}.
Successful training commonly involves altering the LL maximization objective, such as imposing sparsity constraints \citep{NIPS2008_f11bec14,johnson-etal-2007-bayesian} or heuristically estimating the LL value \citep{spitkovsky-etal-2010-viterbi}.
Theses evidence suggests that the LL function might not provide robust information to distinguish between constituents and non-constituents, rendering LL an insufficient objective function for unsupervised parsing.

In this paper, we propose a novel objective for training unsupervised parsers: maximizing SemInfo (the semantic information encoded in constituent structures).
Specifically, we introduce a bag-of-substrings model to represent the sentence semantics with substring statistics, in parallel to how bag-of-words models represent document topics with word statistics.
Next, we estimate the semantic information encoded in substrings (i.e., substring-semantic information) by applying the Probability-Weighted Information (PWI) metric \citep{Aizawa_2003} developed for the bag-of-words model to our bag-of-substrings model.
Finally, we calculate the SemInfo value of a constituent structure by summing up the substring-semantic information associated with the structure.
Experiments show a much stronger correlation between SemInfo and parsing accuracy than the correlation between LL and parsing accuracy.
The improved correlation suggests SemInfo is an effective objective function for unsupervised constituency parsing.
In addition, we develop a Tree Conditional Random Field (TreeCRF)-based model to apply the mean-field SemInfo maximization training to PCFG parsers (the state-of-the-art non-ensemble method for unsupervised constituency parsing \citep{liu-etal-2023-simple}).
Experiments demonstrate that the SemInfo maximization objective improves the PCFG's parsing accuracy by 7.85 sentence-F1 scores across five latest PCFG variants and in four languages.

Our main contributions are: (1) Proposing a novel method for estimating SemInfo, the semantic information encoded in constituent structures. (2) Demonstrating a strong correlation between SemInfo values and parsing accuracy. (3) Developing a TreeCRF model to apply mean-field SemInfo maximization training to PCFG parsers, significantly improving parsing accuracy and achieving state-of-the-art level results as non-ensemble parsers.

\section{Background}
\def\fracvphan{\vphantom{ \left(\frac{a^{0.3}}{b}\right) } }
\label{sec:background}
The idea that constituent structures reflect the organization of sentence semantics is central to modern linguistic studies \citep{Steedman_2000, Pollard_Sag_1987}.
A constituent is a substring $s$ in a sentence $x$ that can function independently \citep{Carnie_2007} and carries self-contained meanings \citep{Heim_Kratzer_1998}.
A collection of constituents forms a tree-shaped structure $t$, which we can represent as a collection of its constituent substrings $t=\{s_1, s_2, ...\}$.
For example, the constituent structure in the top right of Figure~\ref{fig:substring-semantics} can be represented as $\{$``a theory'', ``until late night'',...$\}$.
Previous research \citep{Shen2017NeuralLM, yang-etal-2021-pcfgs} measures the accuracy of the parsing prediction by instance level sentence-F1 ($\text{SF1}^i$) score.
Aggregating the $\text{SF1}^i$ score over the corpus gives the corpus-level sentence-F1 score (\csfo), which previous research used to evaluate the parser quality.

In this paper, we will apply the Probability-Weighted Information (PWI) \citep{Aizawa_2003} designed to measure word-topic information in bag-of-words models (Figure~\ref{fig:bag-of-words}) to measuring substring-semantic information in our bag-of-substrings model.
PWI is an information-theoretic interpretation of the term frequency-inverse document frequency (tf-idf) statistic.
The tf-idf statistic is an effective feature in finding keywords in documents \citep{Li_Fan_Zhang_2007} or in locating documents based on the given keyword \citep{mishra2015analysis}.
Let $\gD$ denote a document corpus, $d_i$ the $i$-th document in the corpus, and $w_{ij}$ the $j$-th word in $d_i$.
The bag-of-words model represents the document $d_i$ as an unordered collection of words occurring in the document (i.e., $d_i=\{w_{i1}, w_{i2},...\}$).
Tf-idf, as shown in Equation~\ref{eq:tfidf}, is the product of the term frequency $F(w_{ij}, d_i)$ (i.e. the frequency of $w_{ij}$ occurring in $d_i$) and the inverse document frequency (i.e. the inverse log-frequency of documents containing $w_{ij}$).
PWI interprets the term frequency as the word generation probability and the inverse document frequency as the piecewise word-document information (Equation~\ref{eq:pwi}).
The PWI value estimates the information that $w_{ij}$ carries with regard to $d_i$.
A high value indicates that $w_{ij}$ is both frequent in $d_i$ and strongly associated with $d_i$.
In other words, $w_{ij}$ is a keyword of $d_i$.
\begin{small}
    \begin{align}
    \text{tf-idf}(w_{ij}, d_{i})&= \underbrace{\fracvphan F(w_{ij}, d_{i})}_{\text{term frequency}}\times \underbrace{\log\frac{|\gD|}{|d^\prime: d^\prime\in\gD \wedge w_{ij}\in d^\prime|}}_{\text{inverse document frequency}}\label{eq:tfidf}\\
    &\approx \underbrace{\fracvphan P(w_{ij}|d_i)}_{\text{word generation probability}} \times \underbrace{\log \frac{P(d_i|w_{ij})}{P(d_i)}}_{\text{piecewise word-document information}}\label{eq:pwi}\\
    &= PWI(w_{ij}, d_i)\notag
\end{align}
\end{small}

Our method is developed upon the finding of \citet{chen-etal-2024-unsupervised}: constituent structures can be predicted by searching for frequent substrings among semantically similar paraphrases.
We extend their findings, interpreting the substring frequency statistic as a dominating term in our proposed substring-semantics information metric and applying it to improve unsupervised PCFG training.
As we will see in Section~\ref{sec:exp-improvements}, our method significantly outperforms theirs in three out of the four languages tested.

PCFG is currently the state-of-the-art non-ensemble model for unsupervised constituency parsing \citep{liu-etal-2023-simple,yang-etal-2021-neural}. 
Previous research trains binary PCFG parsers on a text corpus by maximizing the average LL of the corpus.
PCFG is a generative model defined by a tuple $(NT, T, R, S, \pi)$, where $NT$ is the set of non-terminal symbols, $T$ is the set of terminal symbols, $R$ is the set of production rules, $S$ is the start symbol, and $\pi$ is the probability distribution over the rules.
The generation process starts with the start symbol $S$ and iteratively applies non-terminal expansion rules ($A\rightarrow BC: A, B, C\in NT$) or terminal rewriting rules ($A\rightarrow w: A\in NT, w\in T$) until it produces a complete sentence $x$.
We can represent the generation process with a tree-shaped structure $t$.
The PCFG assigns a probability for each distinct way of generating $x$, defining a distribution $P(x, t)$. 
The Inside-Outside algorithm \citep{Baker_1979} provides an efficient solution for computing the total sentence probability $P(x)=\mathop{\sum}_{t}P(x, t)$.
It constructs a $\beta(s, A)$ table that records the total probability of generating a substring $s$ of $x$ from the non-terminal $A$.
The sentence probability can be calculated as $P(x)=\beta(x, S)$, the probability of $x$ being generated from the start symbol $S$.
The $\beta(x, S)$ quantity is commonly referred to as $Z(X)$ \citep{eisner-2016-inside}.
Besides the total sentence probability, the $\beta$ table can also be used to calculate the span-posterior probability of $s$ being a constituent \citep{eisner-2016-inside} (Equation~\ref{eq:spanposterior}).\footnote{We explain the derivation in more detail in Section~\ref{appendix:span-posterior-by-bp}.}
\begin{small}
    \begin{equation}
    P(s\text{ is a constituent}|x) = \sum_{A\in NT} \frac{\partial \log Z(x)}{\partial \log \beta(s, A)}
    \label{eq:spanposterior}
\end{equation}
\end{small}

Span-based TreeCRF model is widely adopted in constituency parsers \citep{kim-etal-2019-unsupervised, stern-etal-2017-minimal}.
It models the parser distribution $P(t|x)$, the probability of constituent structure $t$ given $x$.
It determines the probability of $t$ by evaluating whether all substrings involved in the structure are constituents.
It assigns a high score to a substring $s$ in its potential function $\phi(s, x)$ if $s$ is likely a constituent and a low score if $s$ is unlikely a constituent.
Subsequently, It can represent the parser distribution as $P(t|x)\propto \prod_{s\in t}\phi(s, x)$.
In previous research, $\phi(s, x)$ has been parameterized differently, such as using the span posterior probability for decoding ($\phi(s, x)=P(s\text{ is a constituent}|x)$) \citep{yang-etal-2021-pcfgs}  or using the exponentiated output from Long-Short Term Memory model ($\phi(s, x)=\exp(LSTM(x, s))$) \citep{kim-etal-2019-unsupervised}.

\section{SemInfo: A Metric of Semantic Information Encoded in Constituent Structures}
\label{sec:seminfo}
In this section, we introduce our estimation method of SemInfo, the semantic information encoded in constituent structures. 
We first propose a bag-of-substrings model (Figure~\ref{fig:paraphraser}), representing the semantics of a sentence by examining how substrings in the sentence are \emph{regenerated} during a paraphrasing process.
We assume the paraphrasing process is capable of generating \emph{natural language paraphrases} (i.e., the paraphrases should both be acceptable as natural language sentences and have similar semantics to the original sentence).
We use instruction-following large language models (LLMs) for the paraphrasing model, exploiting their outstanding zero-shot learning capability \citep{DBLP:journals/corr/abs-2306-04757}.
Next, we apply the PWI metric \citep{Aizawa_2003} to measure the substring-semantics information, utilizing the parallel structure between the bag-of-words model and our bag-of-substrings model (Figure~\ref{fig:parallel-bow-bos}).
Finally, we estimate the SemInfo value for constituent structures by summing the substring-semantics information associated with the structure.

\subsection{Defining Substring-Semantic Information using Bag-of-Substrings Model}
\begin{figure}[t]
    \centering
    \begin{subfigure}{0.49\linewidth}
        \centering
        \includegraphics[width=0.45\linewidth]{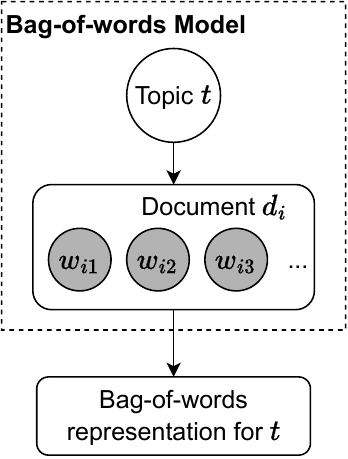}
        \caption{Bag-of-words model.}
        \label{fig:bag-of-words}
    \end{subfigure}
    \begin{subfigure}{0.49\linewidth}
        \centering
        \includegraphics[width=0.7\linewidth]{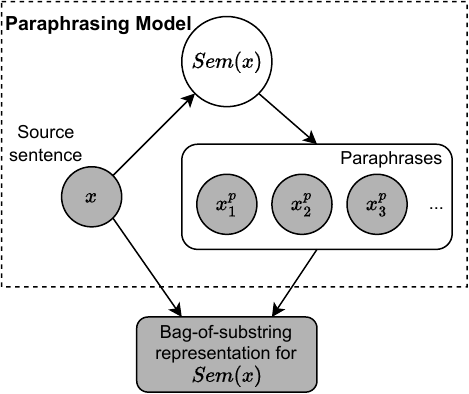}
        \caption{Bag-of-Substrings model.}
        \label{fig:paraphraser}
    \end{subfigure}
    \caption{Parallel structure between the traditional bag-of-words representation of topics and the proposed bag-of-substrings representation of semantics. }
    \label{fig:parallel-bow-bos}
\end{figure}

Our bag-of-substrings model shares a parallel structure with the traditional bag-of-words model.
As discussed in Section~\ref{sec:background}, the bag-of-words model can model the word-topic information using the PWI metric.
Exploiting the structural parallelism, we can apply the PWI metric to our bag-of-substrings model to estimate the information between substrings and sentence semantics (Equation~\ref{eq:substring-semantics-info}).

The bag-of-substrings model is based on the paraphrasing model $P(x^p|Sem(x))$ shown in Figure~\ref{fig:paraphraser}.
The paraphrasing model takes a source sentence $x$ as input, internally analyzes its semantics $Sem(x)$, and generates a paraphrase $x^p$.
We can repeatedly sample from the process, collecting a paraphrase set $\sX^p=\{x^p_1, x^p_2,...\}$.
We define the bag-of-substrings model by examining whether a substring $s$ of $x$ appears in $\sX^p$.
We consider the appearance of $s$ in $\sX^p$ as $s$ being generated by the bag-of-substrings model.
The generation modeling establishes a relationship between the semantics $Sem(x)$ and the substring $s$, which we will use to estimate the substring-semantic information.

The PWI metric requires two components to calculate the substring-semantic information: $P(s|Sem(x))$, the substring generation probability, and $\log\frac{P(Sem(x)|s)}{P(Sem(x))}$, the piecewise mutual information between $s$ and $Sem(x)$.
Similar to the bag-of-words model, we will calculate the two components using the frequency of $s$ in $\sX^p$ and the inverse frequency of $s$ in the corpus $\mathcal{D}$.
\begin{small}
    \begin{equation}
    I(s, Sem(x)) = P(s|Sem(x))\log \frac{P(Sem(x)|s)}{P(Sem(x))}
    \label{eq:substring-semantics-info}
\end{equation}

\end{small}

\subsection{Calculating PWI using Maximal Substrings}

\begin{wrapfigure}{r}{0.5\textwidth}
    \centering
    \includegraphics[width=\linewidth]{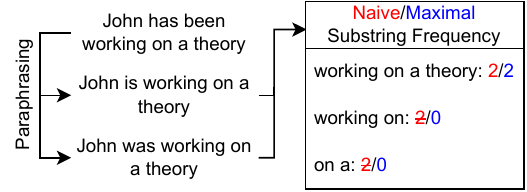}
    \caption{An example for naive substring frequency among paraphrases failing to estimate $P(s|Sem(x))$.}
    \label{fig:simple-freq}
\end{wrapfigure}

Naively measuring substring frequency among paraphrases $\sX^p$ will yield a misleading estimate of $P(s|Sem(x))$.
The reason is that one substring can be nested in another substring.
If a substring $s$ is generated to convey semantic information, we will observe an occurrence of $s$ along with an occurrence of all its substrings.
Hence, the naive substring frequency will wrongly count substring occurrences caused by the generation of larger substrings as occurrences caused by $P(s|Sem(x))$.
Let us consider the example illustrated in Figure~\ref{fig:simple-freq}.
All three substrings in the example have a frequency of 2, yet only the first substring carries significant semantic information.
This is because the occurrence of the first substring causes the occurrence of the second and third substrings.
The true frequency of the second and third substrings should be 0 instead of 2.

We introduce the notion of maximal substring to counter this problem. 
Given a source sentence $x$ and a paraphrase $x^p_i$, the maximal substring between the two is defined in Equation~\ref{eq:maximal-substr}.
Intuitively, a maximal substring is the largest substring that occurs in both $x$ and $x^p_i$.
Formally, we denote the partial order relationship of string $\alpha$ being a substring in string $\beta$ by $\alpha\leq\beta$, and denote the set of maximal substrings by $MS(x, x^p_i)$.
Using maximal substrings, we can avoid over-counting substring occurrences caused by the generation of larger substrings.
\begin{small}
    \begin{equation}
    MS(x, x^p_i):=\{\alpha: \alpha\leq x\wedge \alpha\leq x^p_i\wedge \forall \alpha^\prime (\alpha<\alpha^\prime \implies \neg \alpha^\prime \leq x \vee \neg \alpha^\prime \leq x^p_i) \}
    \label{eq:maximal-substr}
\end{equation}
\end{small}

% ???NEEDS TO CHANGE NARRATIVE HERE
We are now ready to define $P(s|Sem(x))$ using the paraphrasing distribution $P(x^p|Sem(x))$ and the notion of maximal substrings.
We define $P(s|Sem(x))$ to be proportional to $s$'s probability of being generated as a maximal substring in paraphrases (Equation~\ref{eq:word-gen-prob-as-expected-maximal-substring-prob}).
The probability can then be approximated using the maximal substring frequency $F(s, \sX^p)$, as shown in Equation~\ref{eq:word-gen-prob-as-freq}.
\begin{small}
    \begin{align}
    P(s|Sem(x))&\propto \mathop{\E}_{x_i^p\sim P(x^p|Sem(x))} \1(s\in MS(x_i^p, x))\label{eq:word-gen-prob-as-expected-maximal-substring-prob}\\
    &\approx F(s, \sX^p)\label{eq:word-gen-prob-as-freq}
    \end{align}
\end{small}

Similarly, we define the inverse document frequency for maximal substrings (Equation~\ref{eq:inverse-dfreq}). 
The inverse document frequency can serve as an estimate of the piecewise substring-semantics information, quantifying how useful a substring is to convey semantic information.
A high inverse document frequency implies that only a few $Sem(x)$ in the corpus generate $s$ as their maximal substring. 
In other words, we can easily identify the target semantics by examining whether $s$ appears as maximal substrings.
\begin{small}
    \begin{align}
    \log \frac{P(Sem(x)|s)}{P(Sem(x))}&\approx\log\frac{|\gD|}{|\{x^\prime: x^\prime\in\gD \wedge  s\in MS(x, x^\prime)\}|}\label{eq:inverse-dfreq}
\end{align}
\end{small}

\subsection{Estimating SemInfo}
A constituent structure $t$ can be represented as a set of constituent substrings. 
We define SemInfo, the information between $t$ and $Sem(x)$, as the cumulative substring-semantics information associated with $t$ (Equation~\ref{eq:seminfo}).
We estimate the substring-semantics information with the maximal substring frequency-inverse document frequency developed in the above section.
\begin{small}
    \begin{align}
    I(t, Sem(x)) &= \sum_{s\in t} I(s, Sem(x))\label{eq:seminfo}\\
    &\propto \sum_{s\in t}\underbrace{F(s, \sX^p)\log\frac{|\gD|}{|\{x^\prime: x^\prime\in\gD \wedge  s\in MS(x, x^\prime)\}|}}_{\text{maximal substring frequency-inverse document frequency}}
\end{align}
\end{small}

\section{SemInfo Maximization via TreeCRF Model}
\label{sec:treecrf} 
\begin{figure}[t]
    \centering
    % \vspace{-0.8cm}
    \includegraphics[width=0.9\linewidth]{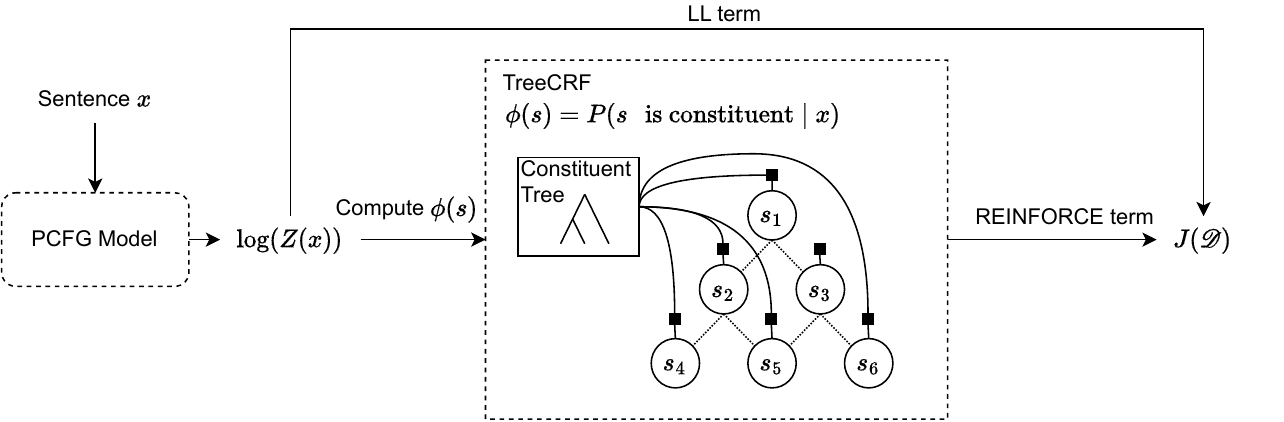}
    \caption{Pipeline of our SemInfo maximization training}
    % \vspace{-0.6cm} 
    \label{fig:pipeline} 
\end{figure}  

We train our PCFG models on Equation~\ref{eq:loss} using the pipeline shown in Figure~\ref{fig:pipeline}.
The pipeline consists of three steps:
(1) We compute the $\log (Z(x))$ by applying the inside algorithm on the PCFG model.
This step yields the leading log-likelihood term in Equation~\ref{eq:loss}.
More importantly, it constructs the computation graph needed to calculate the span-posterior probability $P(s\text{ is a constituent}|x)$.
(2) We extract the span-posterior probability via back-propagating $\log (Z(x))$ and parameterize a TreeCRF model by setting $\phi(s, x) = P(s\text{ is a constituent}|x)$ (Equation~\ref{eq:TreeCRF-potential}).
This parametrization leads to a tree distribution $P^{CRF}(t|x)$ that functions as an \emph{one-step Reinforcement Learning agent}.
(3) We train the PCFG model by applying the SemInfo maximization on $P^{CRF}(t|x)$.
This TreeCRF-based training method is equivalent to applying a mean-field SemInfo maximization to the PCFG model.
We choose the TreeCRF model because it enables efficient sampling from $P^{CRF}(t|x)$ and entropy calculation for the distribution.
As discussed in Appendix~\ref{appendix:asm}, the TreeCRF-based training method performs equivalently to applying SemInfo maximization on the PCFG model directly. 
Yet, the TreeCRF-based method runs 5x faster and uses $\frac{1}{6}$ the memory compared to the direct PCFG optimization method.

We apply the REINFORCE algorithm with average baseline \citep{DBLP:journals/ml/Williams92} to facilitate the training.
We include the maximum entropy regularization \citep{Ziebart2008MaximumEI} and the traditional LL term $\log Z(x)$ in the training.
Notably, the LL term significantly stabilizes the training process.
This stabilization effect may be related to the strong correlation between LL values and parsing accuracy at the early training stage, as discussed in Section~\ref{sec:corpus-level-correlation}.
\begin{small}
    \begin{align}
            P^{CRF}(t|x) &\propto \prod_{s\in t}P(s\text{ is a constituent}|x) \notag \\
            &=\prod_{s\in t} \sum_{A\in NT}\frac{\partial\log Z(x)}{\partial \log \beta(s, A)}
            \label{eq:TreeCRF-potential}
    \end{align}
\end{small}

\begin{small}
    \begin{equation}
        \begin{aligned}
            \gJ(\gD) =  \mathop{\E}_{x\sim \mathcal{D}}[\log Z(x) + \mathop{\E}_{t\sim P^{CRF}(t|x)}[&\log P^{CRF}(t|x) (I(t, Sem(x)) -\\
    &\mathop{\E}_{t\sim P^{CRF}(t|x)}I(t, Sem(x)) + \beta H(P(t|x)))]]\label{eq:loss}
        \end{aligned}
    \end{equation}
    
\end{small}

\section{Experiment}

\subsection{Experiment Setup}
\label{sec:exp-setup}
We evaluate the effect of the SemInfo maximization objective on five latest PCFG variants: Neural-PCFG (NPCFG), Compound-PCFG (CPCFG) \citep{kim-etal-2019-compound}, TNPCFG \citep{yang-etal-2021-pcfgs}, Simple-NPCFG (SNPCFG), and Simple-CPCFG (SCPCFG) \citep{liu-etal-2023-simple}.\footnote{Our implementation is based on the source code of \citet{yang-etal-2021-pcfgs} and \citet{liu-etal-2023-simple}}
SNPCFG and SCPCFG represent the current state-of-the-art for non-ensemble unsupervised constituency parsing.
We use 60 NTs for NPCFG and CPCFG, and 1024 NTs for TNPCFG, SNPCFG, and SCPCFG in our experiment. 
We conduct the evaluations in three datasets and four languages, namely Penn TreeBank (PTB) \citep{Treebank-3} for English, Chinese Treebank 5.1 (CTB) \citep{CTB5.1} for Chinese, and SPMRL \citep{seddah-etal-2013-overview} for German and French.
We adopt the standard data split for the PTB dataset (Sections 02-21 for training, Section 22 for validation, and Section 23 for testing) \citep{kim-etal-2019-compound}.
We adopt the official data split for the CTB and SPMRL datasets. 

Following \citet{Shen2017NeuralLM}, we train the PCFG model on raw text without punctuations and evaluate its parsing performance using the \csfo\nspace scores.
When computing the \csfo\nspace score, we aggregate \isfo\nspace only for sentences longer than two words and drop trivial spans (i.e., sentence-level spans and spans with only one word). 
We use both the \csfo\nspace and \isfo\nspace scores to evaluate the correlation between the SemInfo value and parsing accuracy.

We use the \texttt{gpt-4o-mini-2024-07-18} model as our paraphrasing model and apply the same word normalization techniques as in \citet{chen-etal-2024-unsupervised}.
The average paraphrasing cost is about 5 USD using OpenAI's batch API. 
We use eight semantic-preserving prompts for the paraphrasing model.\footnote{Detailed prompts are listed in Section~\ref{appendix:prompts}}
We apply the snowball stemmer \citep{bird-loper-2004-nltk} to normalize the source sentence and its paraphrases before calculating the maximal substring frequency and the inverse document frequency.
We apply the log-normalization \citep{Sparck_Jones_1972} to the maximal substring frequency to avoid some high-frequency substrings dominating the SemInfo value.
The log-normalization is compatible with the PWI framework, which treats the normalization as an optional step to estimate $P(s|Sem(x))$.
In preliminary experiments, the log-normalization variant performs marginally but consistently better than the unnormalized variant.

\subsection{SemInfo Maximization Significantly Improves Parsing Accuracy}
\label{sec:exp-improvements}
\begin{table}[t]
\vspace{-0.5cm}
    \centering
    \begin{adjustbox}{width=\textwidth}
    \begin{tabular}{l|cc|cc|cc|cc}
    \hline
         & \multicolumn{2}{|c|}{English}  & \multicolumn{2}{c|}{Chinese}   & \multicolumn{2}{c|}{French}  & \multicolumn{2}{c}{German}  \\ \cline{2-9}
        ~ & SemInfo (Ours) & LL & SemInfo & LL & SemInfo & LL & SemInfo & LL \\ \hline
        CPCFG & \textbf{65.74}\textsubscript{±0.81} & 53.75\textsubscript{±0.81} & 50.39\textsubscript{±0.87} & 51.45\textsubscript{±0.49} & \textbf{52.15}\textsubscript{±0.75} & 47.50\textsubscript{±0.41} & \textbf{49.80}\textsubscript{±0.31} & 45.64\textsubscript{±0.73} \\ 
        NPCFG & \textbf{64.45}\textsubscript{±1.13} & 50.96\textsubscript{±1.82} & \textbf{53.30}\textsubscript{±0.42} & 42.12\textsubscript{±3.07} & \textbf{52.36}\textsubscript{±0.62} & 47.95\textsubscript{±0.09} & \textbf{50.74}\textsubscript{±0.28} & 45.85\textsubscript{±0.63} \\ 
        SCPCFG & \textbf{67.27}\textsubscript{±1.08} & 49.42\textsubscript{±2.42} & 51.76\textsubscript{±0.54} & 46.20\textsubscript{±3.65} & \textbf{52.79}\textsubscript{±0.80} & 45.03\textsubscript{±0.42} & \textbf{47.97}\textsubscript{±0.76} & 45.50\textsubscript{±0.71} \\ 
        SNPCFG & \textbf{67.15}\textsubscript{±0.62} & 58.19\textsubscript{±1.13} & \textbf{51.55}\textsubscript{±0.82} & 43.79\textsubscript{±0.39} & \textbf{55.21}\textsubscript{±0.47} & 49.64\textsubscript{±0.91} & \textbf{49.65}\textsubscript{±0.29} & 40.51\textsubscript{±1.26} \\ 
        TNPCFG &\textbf{66.55}\textsubscript{±0.96} & 53.37\textsubscript{±4.28} & 51.79\textsubscript{±0.83} & 45.14\textsubscript{±3.05} & \textbf{54.11}\textsubscript{±0.66} & 39.97\textsubscript{±4.10} & \textbf{49.26}\textsubscript{±0.64} & 44.94\textsubscript{±1.34} \\ \hline
        Average $\Delta$ & \multicolumn{2}{|c|}{+13.09} & \multicolumn{2}{|c|}{+6.02} & \multicolumn{2}{|c|}{+7.31}  & \multicolumn{2}{|c}{+4.92} \\ \hline\hline
        MaxTreeDecoding & \multicolumn{2}{|c|}{58.28}  & \multicolumn{2}{|c|}{49.03} & \multicolumn{2}{|c|}{52.03} & \multicolumn{2}{|c}{50.82} \\ 
        GPT4o-mini & \multicolumn{2}{|c|}{36.16}  & \multicolumn{2}{|c|}{11.82} & \multicolumn{2}{|c|}{30.01} & \multicolumn{2}{|c}{33.56} \\ \hline
    \end{tabular}
    \end{adjustbox}
    \caption{\csfo\space scores of five PCFG variants trained with SemInfo and LL. Each cell in the upper section reports the mean \csfo score and the standard deviation across three \emph{identical and independently trained} PCFG models. Average $\Delta$ indicates average improvements in the \csfo score when training with SemInfo compared to LL. Improvements that are statistically significant ($p<0.05$) are highlighted in bold.}
    \label{tbl:val}

\vspace{-0.5cm}
    
\end{table}
Table~\ref{tbl:val} compares SemInfo-trained PCFGs and LL-trained PCFGs on five contemporary PCFG variants and four languages.
For each variant, we independently train three PCFG models on the SemInfo and LL objectives and report the mean and standard deviation of their \csfo scores.
We can observe that most SemInfo-trained PCFGs achieve significantly higher parsing accuracy than their LL-trained counterparts.
The average improvements are 13.09, 6.02, 7.31, and 4.92 \csfo\nspace scores in English, Chinese, French, and German, respectively. 
Two-tailed t-tests indicate the improvement to be statistically significant (p$<$0.05) in 17 out of 20 combinations.
Two of the three insignificant results are due to the high score variance of the LL-trained PCFGs. 
The significant improvement demonstrates the benefit of the SemInfo maximization objective in the unsupervised constituency parsing task.
The result also confirms the importance of semantic factors in identifying the syntactic constituent structure.

Table~\ref{tbl:val} also compares the SemInfo trained PCFG with two baseline parsers: Maximum Tree Decoding (MTD) parser, which predicts the structure with maximum SemInfo value, and GPT4o-mini parser that asks the GPT4o-mini model to predict the structure in bracket form directly. 
Among the two baselines, we see that the MTD parser has significantly higher \csfo\nspace scores than the GPT4o-mini parser across the four languages.
The accuracy gap indicates that SemInfo is discovering non-trivial information about the constituent structure.
Comparing the SemInfo-trained PCFG and the MTD parser, we see that all SemInfo-trained PCFG variants outperform the MTD parser in English, Chinese, and French. 
The accuracy improvement indicates that the constituent information provided by the SemInfo value is noisy, and the grammar learns to mitigate the noises.
We can again confirm PCFG's de-noising effect in an experiment investigating how paraphrasing noise affects parsing performance (Appendix~\ref{appendix:robust-paraphrasing-noise}). 

In German, SemInfo-trained PCFGs perform worse than the MTD parser.
One possible reason is that the German validation/testing set has a significantly different word vocabulary compared to the training set, unlike the datasets in the other three languages.
The out-of-vocabulary rate in the German dataset is 14\%, while the rate is 5\%, 6\%, and 7\% in the English, Chinese, and French datasets.
This shift in word distribution might be a significant factor in German PCFGs' poor parsing accuracy.

\subsection{SemInfo Strongly Correlates with Parsing Accuracy}
\label{sec:seminfo-corr}
In this section, we investigate how the SemInfo and LL functions contribute to obtaining high-quality PCFG parsers from two aspects: (1) Whether the function can accurately evaluate the model's prediction (measured by \isfo). (2) Whether the function can approximately rank PCFG parsers in accordance with parsing performance (measured by \csfo).
Our experiments indicate that SemInfo can serve as an accurate estimate of parsing accuracy and that SemInfo is a better training objective for unsupervised parsers than LL.
We evaluate the two aspects using the Spearman correlation \citep{Spearman_1904} between the SemInfo/LL values and the \isfo/\csfo scores. 
We refer to the correlation analysis using the \isfo\nspace score \textit{sentence-level} analysis and the analysis using the \csfo\nspace score \textit{corpus-level} analysis.

\subsubsection{SemInfo Estimates Parsing Accuracy}
The sentence-level analysis assesses the SemInfo/LL's capability to evaluate the model prediction accurately.
We independently train eight \emph{identical} PCFG models using the LL maximization objective.
Each model is trained with a unique random seed for 30k steps.
These eight models produce eight (\isfo, SemInfo, LL) tuples for any given sentence, which we use to calculate the sentence-level Spearman correlation coefficient.

Figure~\ref{fig:corr-example} illustrates the correlation gap between SemInfo-\isfo\nspace and LL-\isfo\nspace pairs using two sentences in the English validation set.
Between the two sentences, the SemInfo-\isfo\nspace pairs exhibit positive correlations while the LL-\isfo\nspace pairs exhibit no apparent correlations.
Table~\ref{tbl:corr} confirms the correlation gap using the correlation coefficient aggregated in the corpus level. 
We perform mean-aggregation using Fisher's Z transformation \citep{Fisher_1915}.
The transformation converts the coefficient to a uni-variance distribution and reduces the negative impact of the aggregation caused by the coefficient's skewed distribution \citep{Silver_Dunlap_1987}.
In the table, we observe that the aggregated coefficients for the SemInfo-\isfo\nspace pairs range from 0.6-0.9, whereas the aggregated coefficients for the LL-\isfo\nspace correlation center around 0.
We can also consistently observe the correlation gap across multiple training stages, as further discussed in Appendix~\ref{appendix:sent-corr-stages}.
The consistent correlation gap, on the one hand, suggests that SemInfo can serve as an accurate estimate of parsing accuracy and that SemInfo is a better training objective for unsupervised parsers than LL. 
On the other hand, it highlights SemInfo's ability to capture constituent information, reaffirming a close relationship between constituent structure and sentence semantics.

\begin{table}
% \vspace{-0.6cm}
\makebox[\textwidth][c]{
    \begin{minipage}[b]{0.45\textwidth}
    \centering
    \begin{adjustbox}{width=\textwidth}
    \begin{tabular}{l|c|c|c}
    \hline
         & SemInfo-\isfo & LL-\isfo & SemInfo-LL \\ \hline
        CPCFG & 0.6518 & 0.0223 & 0.0196 \\ 
        NPCFG & 0.6347 & -0.0074 & -0.0045 \\ 
        SCPCFG & 0.6431 & -0.0013 & 0.0505 \\ 
        SNPCFG & 0.9289 & 0.0102 & 0.0182 \\ 
        TNPCFG & 0.6449 & 0.1077 & 0.1426 \\ \hline
    \end{tabular}
    \end{adjustbox}
    \captionof{table}{Spearman correlation coefficient among (SemInfo, LL, \isfo), and LL over the English validation set. Correlations are aggregated at the corpus-level.}
    \label{tbl:corr}
    \end{minipage}
    \hspace{0.02\textwidth}

    \begin{minipage}[b]{0.53\textwidth}
        \centering
        \includegraphics[width=\linewidth]{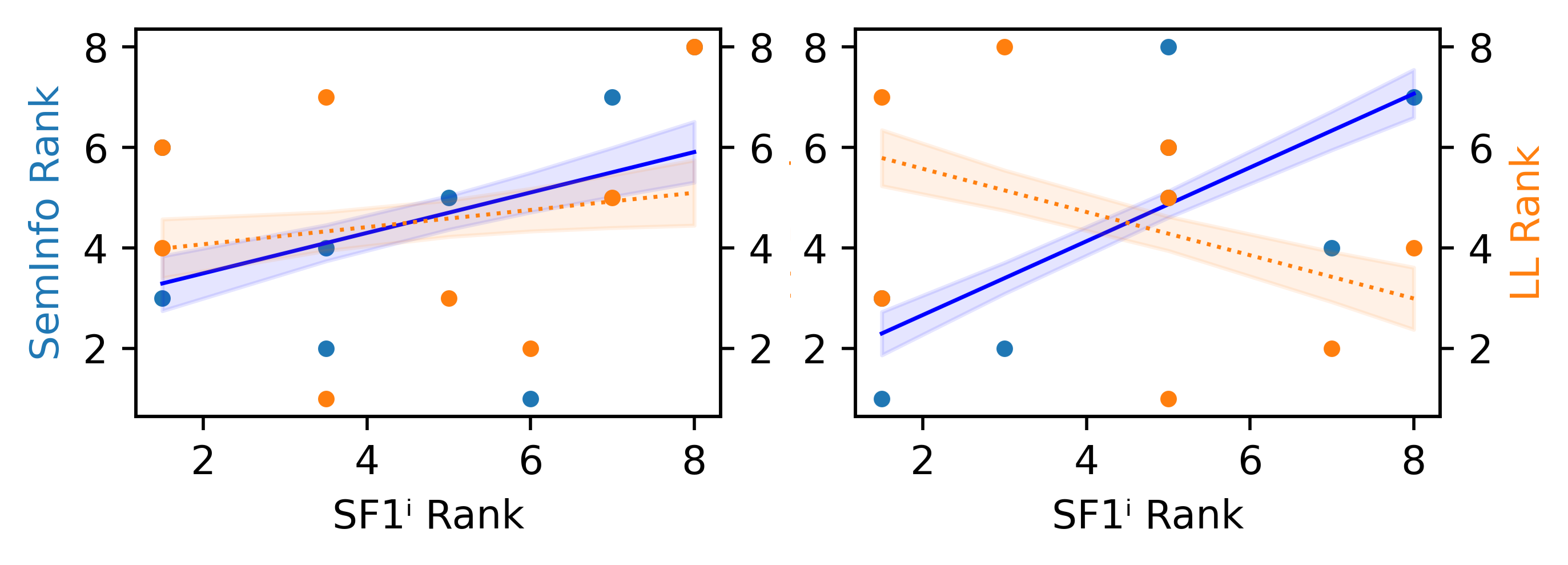}
        % \vspace{-0.3cm}
        \captionof{figure}{Spearman rank analysis of (SemInfo, LL, \isfo) pairs obtained from eight independently trained NPCFG models. The values are measured on two sentences in the English dataset. Please refer to Figure~\ref{fig:sent_level-corr-6figs} for more examples.}
        \label{fig:corr-example} 
    \end{minipage}
}
% \vspace{-0.8cm}
\end{table}

\subsubsection{SemInfo Ranks PCFG Models Better than LL}
\label{sec:corpus-level-correlation}
The corpus-level analysis evaluates the SemInfo/LL's capability to rank PCFG parsers by their parsing performance.
We examine the correlation using model checkpoints collected over different training stages of the above eight PCFG models. 
Each stage is represented by a window over the amount of training steps.
For example, a stage [1k, 10k] contains checkpoints from 1k to 10k steps. 
These checkpoints produce a set of (\csfo, corpus-averaged SemInfo, corpus-averaged LL) tuples, which we use to calculate the corpus-level coefficient at that training stage.

\begin{wrapfigure}{r}{0.45\textwidth}
    \includegraphics[width=\linewidth]{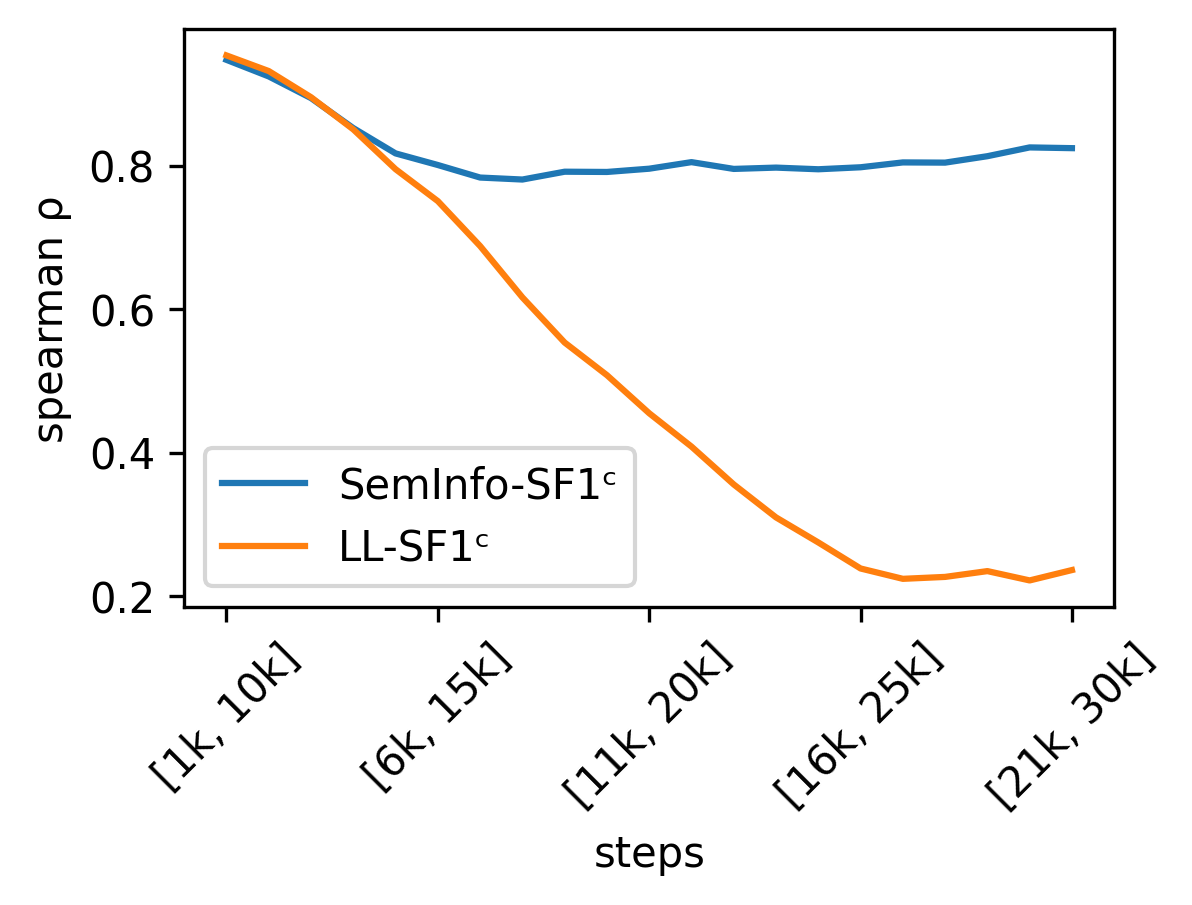}
    \caption{Spearman $\rho$ with \csfo\nspace in different training stages of NPCFG.}
    \label{fig:corpus-corr}
\end{wrapfigure}
Figure~\ref{fig:corpus-corr} illustrates the SemInfo-\csfo\nspace and LL-\csfo\nspace correlation curves for NPCFG.\footnote{We include the correlation curve for the other four PCFG variants in Appendix~\ref{appendix:corpus-corr}.}
We can observe that LL does have a strong corpus-level correlation with \csfo\nspace at the early stage of training despite having a near-non-existent sentence-level correlation. 
However, LL's coefficient quickly diminishes as training progresses, dropping below 0.4 at the late training stage. 
This result indicates that LL identifies a reasonable PCFG parser among a set of poorly performing parsers in the early training stage, explaining why the LL-training can result in non-trivial PCFG parsers despite having negligible correlation in the sentence-level analysis.
Yet, this ability quickly degrades as the training progresses.
In comparison, SemInfo maintains a strong correlation across the whole training process, which indicates SemInfo's superior capability in ranking PCFG parsers by their performance.

\subsection{Comparing with State-Of-The-Arts}

\begin{table}[t]
% \vspace{-0.4cm}
    \centering
    \footnotesize
    % \begin{adjustbox}{width=0.8\textwidth}
        \begin{tabular}{l|c|c|c|c}
        \hline
            & English & Chinese & French & German \\ \hline
            % CPCFG (60NT) & 64.89\textsubscript{±0.88} & 51.03\textsubscript{±0.97} & 51.43\textsubscript{±0.50} & 47.06\textsubscript{±0.33} \\ \hline
            NPCFG (60NT) & 63.62\textsubscript{±1.07} & \textbf{53.92}\textsubscript{±0.48} & 51.88\textsubscript{±0.73} & 47.77\textsubscript{±0.26} \\ 
            SCPCFG (1024NT) & \textbf{66.92}\textsubscript{±0.76} & 52.26\textsubscript{±0.41} & 52.29\textsubscript{±0.53} & 45.32\textsubscript{±0.67} \\ 
            SNPCFG (1024NT) & 66.84\textsubscript{±0.53} & 52.04\textsubscript{±0.93} & \textbf{54.37}\textsubscript{±0.10} & 47.27\textsubscript{±0.16} \\ \hline
            % TNPCFG (1024NT) & 66.01\textsubscript{±0.83} & 52.41\textsubscript{±0.85} & 53.18\textsubscript{±0.39} & 46.33\textsubscript{±0.69} \\ \hline
            \hline
            % % ~ & ~ & ~ & ~ & ~ \\ \hline
            % GPT4o mini & 37.96 & 11.63 & 31.13 & 31.2 \\ \hline
            % MaxDecoding & 58.25 & 49.2 & 51.18 & 49.19 \\ \hline\hline
            % ~ & ~ & ~ & ~ & ~ \\ \hline
            Spanoverlap \citep{chen-etal-2024-unsupervised} & 52.9 & 48.7 & 48.5 & \textbf{49.5} \\
            SCPCFG (2048NT) \citep{liu-etal-2023-simple} & 60.6 & 42.9 & 49.9 & 49.1 \\ 
            SNPCFG (4096NT) \citep{liu-etal-2023-simple} & 65.1 & 39.9 & 38 & 46.7 \\ 
            % NPCFG & 52.3 & 26.3 & 45 & 42.3 \\ \hline
            % CPCFG & 56.3 & 38.7 & 45 & 43.5 \\ \hline
            % TNPCFG (500NT) \citep{yang-etal-2021-pcfgs} & 57.7 & 39.2 & 39.1 & 47.1 \\ \hline
            URNNG \citep{kim-etal-2019-unsupervised}& 40.7 & 29.1 & - & - \\
            NBL-PCFG \citep{yang-etal-2021-neural} & 60.4 & - & - & - \\
            S-DIORA \citep{xu-etal-2021-improved} & 57.6 & - & - & - \\ 
            % StructFormer \citep{shen-etal-2021-structformer} & 54 & - & - & - \\ 
            Constituency Test \citep{cao-etal-2020-unsupervised} & 62.8 & - & - & - \\ \hline
        \end{tabular}
    % \end{adjustbox}
    \caption{\csfo on English, Chinese, French, and German test sets. The top section shows the score for SemInfo-trained PCFGs while the bottom section shows the result from previous work. }
    \label{tbl:test}
    \vspace{-0.4cm}

\end{table}
Table~\ref{tbl:test} compares three SemInfo-trained PCFG variants with the state-of-the-art non-ensemble methods for unsupervised constituency parsing. 
The SemInfo-trained PCFGs achieved state-of-the-art level parsing accuracy in English, Chinese, and French, outperforming the second-best algorithm by 1.82, 11.02, and 4.47 \csfo scores, respectively. 
The SemInfo-trained PCFGs, while using less than half the parameters, perform on par or significantly better than the larger SCPCFG and SNPCFG reported by \citet{liu-etal-2023-simple}.
The comparison showcases the strong parsing accuracy of the SemInfo-trained PCFGs, confirming the usefulness of semantic information in discovering the constituent structure.

\section{Related works}
\textbf{Parsing with PCFG}
Unsupervised PCFG training is a long-established \citep{klein-manning-2002-generative} and state-of-the-art \citep{liu-etal-2023-simple} approach for non-ensemble unsupervised constituency parsing. 
Much research has been dedicated to improving PCFG training from the model perspective, such as scaling up the PCFG model \citep{yang-etal-2021-pcfgs, liu-etal-2023-simple}, integrating lexical information \citep{yang-etal-2021-neural}, and allowing PCFG rule probabilities to condition on sentence embeddings through variational inference \citep{kim-etal-2019-compound}.
Our improvement is from the model optimization perspective and can be combined with the above efforts.
Our experiments validate the effectiveness of the SemInfo maximization objective in improving unlexicalized PCFGs.
The SemInfo maximization objective is also applicable to lexicalized PCFGs, which we leave to future work.

\textbf{Parsing with Semantics}
\citet{zhao-titov-2020-visually} and \citet{zhang-etal-2021-video} have sought to improve PCFG training by learning to identify visual features, maximizing the association between constituent structures and these visual features.
If we consider the visual features as semantic representations, their approach is effectively maximizing the semantic information of the constituent structure.
In comparison, our method shares the same underlying principle but represents the semantics with textual features.
Our method leverages large language models as semantic processors, utilizing their outstanding semantic processing capabilities \citep{Minaee2024LargeLM}. 
We believe that combining both textual and visual semantic representations presents a significant research direction for unsupervised parsing tasks. 

\textbf{Improving Parsing with Ensemble Models} 
Ensembling unsupervised parsers \citep{DBLP:conf/iclr/ShayeghC0CM24} significantly improves accuracy for unsupervised parsing by aggregating predictions from various base parsers. 
They show that those base parsers predict the constituent structure differently and utilize the difference to obtain a more accurate parsing result. 
Our method can be combined with the ensemble method for better parsing accuracy.
We conduct a parser agreement analysis in Appendix~\ref{sec:appendix-ensemble} to show the potential.
The agreement analysis shows an agreement score of 80 among our SemInfo-trained PCFG parsers using various paraphrasing models.
The agreement score is similar to that of homogeneous parsers reported in \citet{DBLP:conf/iclr/ShayeghC0CM24}.
The analysis also shows that our parsers have an agreement score of 50 with other base parsers, similar to the reported score between heterogeneous parsers. 
The similarity in agreement score suggests that our parsers should be able to serve as a useful component in the ensemble method.

\section{Conclusion}
In this paper, we proposed and validated SemInfo maximization as a novel objective for unsupervised constituency parsing. 
We developed a bag-of-substrings model to represent the sentence semantics and applied the probability-weighted information metric to estimate the SemInfo.
We applied the SemInfo maximization objective to training PCFG parsers.
Experiments showed that SemInfo has a strong sentence-level correlation with parsing accuracy and that SemInfo maintains a consistent corpus-level correlation throughout the PCFG training process.
These correlation analyses indicate that SemInfo is an accurate estimate of parsing accuracy and that it is a reliable training objective for unsupervised parsers.
As a result, SemInfo-trained PCFGs significantly outperformed LL-trained PCFGs across four languages, achieving state-of-the-art level performance in three of them.
Our findings highlight the effectiveness of leveraging semantic information in unsupervised constituency parsing, paving the way for semantically-informed unsupervised parsing methods.

\section{Reproducibility}
We provide detailed explanation of our method in Sections~\ref{sec:seminfo} and \ref{sec:treecrf}. 
We outline further implementation details, such as the data source, model architecture, and hyper-parameter settings, in Section~\ref{sec:exp-setup}.
We release the source code at \href{https://github.com/junjiechen-chris/Improving-Unsupervised-Constituency-Parsing-via-Maximizing-Semantic-Information.git}{https://github.com/junjiechen-chris/Improving-Unsupervised-Constituency-Parsing-via-Maximizing-Semantic-Information.git}.

\section{Acknowledgment}
This research was funded by the Japan Society for the Promotion of Science through the Research Fellowships for Young Scientists (Grant No. JP23KJ0565) and by the KAKEN project (Grant No. 24H00087).
We sincerely thank them for their financial support of the research.
We also appreciate the reviewer's thorough evaluation and valuable suggestions during the review process.

\bibliography{src/iclr2025_conference,src/anthology_0, src/anthology_1, src/custom, src/anthology_2}

\begin{thebibliography}{44}
\providecommand{\natexlab}[1]{#1}
\providecommand{\url}[1]{\texttt{#1}}
\expandafter\ifx\csname urlstyle\endcsname\relax
  \providecommand{\doi}[1]{doi: #1}\else
  \providecommand{\doi}{doi: \begingroup \urlstyle{rm}\Url}\fi

\bibitem[Aizawa(2003)]{Aizawa_2003}
Akiko Aizawa.
\newblock An information-theoretic perspective of tf–idf measures.
\newblock \emph{Information Processing \& Management}, 39\penalty0 (1):\penalty0 45–65, January 2003.
\newblock ISSN 03064573.
\newblock \doi{10.1016/S0306-4573(02)00021-3}.
\newblock URL \url{https://linkinghub.elsevier.com/retrieve/pii/S0306457302000213}.

\bibitem[Baker(1979)]{Baker_1979}
J.~K. Baker.
\newblock Trainable grammars for speech recognition.
\newblock \emph{The Journal of the Acoustical Society of America}, 65\penalty0 (S1):\penalty0 S132–S132, June 1979.
\newblock ISSN 0001-4966, 1520-8524.
\newblock \doi{10.1121/1.2017061}.
\newblock URL \url{https://pubs.aip.org/jasa/article/65/S1/S132/739840/Trainable-grammars-for-speech-recognition}.

\bibitem[Bird \& Loper(2004)Bird and Loper]{bird-loper-2004-nltk}
Steven Bird and Edward Loper.
\newblock {NLTK}: The natural language toolkit.
\newblock In \emph{Proceedings of the {ACL} Interactive Poster and Demonstration Sessions}, pp.\  214--217, Barcelona, Spain, July 2004.
\newblock URL \url{P04-3031}.

\bibitem[Cao et~al.(2020)Cao, Kitaev, and Klein]{cao-etal-2020-unsupervised}
Steven Cao, Nikita Kitaev, and Dan Klein.
\newblock Unsupervised parsing via constituency tests.
\newblock In Bonnie Webber, Trevor Cohn, Yulan He, and Yang Liu (eds.), \emph{Proceedings of the 2020 Conference on Empirical Methods in Natural Language Processing (EMNLP)}, pp.\  4798--4808, Online, November 2020. Association for Computational Linguistics.
\newblock \doi{10.18653/v1/2020.emnlp-main.389}.
\newblock URL \url{https://aclanthology.org/2020.emnlp-main.389}.

\bibitem[Carnie(2007)]{Carnie_2007}
Andrew Carnie.
\newblock \emph{Syntax: a generative introduction}.
\newblock Introducing linguistics. Blackwell Pub, Malden, MA, 2nd ed edition, 2007.
\newblock ISBN 9781405133845.

\bibitem[Carroll \& Charniak(1992)Carroll and Charniak]{Carroll1992TwoEO}
Glenn Carroll and Eugene Charniak.
\newblock Two experiments on learning probabilistic dependency grammars from corpora.
\newblock Technical report, Brown University, USA, 1992.

\bibitem[Chen et~al.(2022)Chen, He, and Miyao]{chen-etal-2022-modeling}
Junjie Chen, Xiangheng He, and Yusuke Miyao.
\newblock Modeling syntactic-semantic dependency correlations in semantic role labeling using mixture models.
\newblock In Smaranda Muresan, Preslav Nakov, and Aline Villavicencio (eds.), \emph{Proceedings of the 60th Annual Meeting of the Association for Computational Linguistics (Volume 1: Long Papers)}, pp.\  7959--7969, Dublin, Ireland, May 2022. Association for Computational Linguistics.
\newblock \doi{10.18653/v1/2022.acl-long.548}.
\newblock URL \url{https://aclanthology.org/2022.acl-long.548}.

\bibitem[Chen et~al.(2024)Chen, He, Bollegala, and Miyao]{chen-etal-2024-unsupervised}
Junjie Chen, Xiangheng He, Danushka Bollegala, and Yusuke Miyao.
\newblock Unsupervised parsing by searching for frequent word sequences among sentences with equivalent predicate-argument structures.
\newblock In Lun-Wei Ku, Andre Martins, and Vivek Srikumar (eds.), \emph{Findings of the Association for Computational Linguistics ACL 2024}, pp.\  3760--3772, Bangkok, Thailand and virtual meeting, August 2024. Association for Computational Linguistics.
\newblock URL \url{https://aclanthology.org/2024.findings-acl.225}.

\bibitem[Chia et~al.(2023)Chia, Hong, Bing, and Poria]{DBLP:journals/corr/abs-2306-04757}
Yew~Ken Chia, Pengfei Hong, Lidong Bing, and Soujanya Poria.
\newblock {INSTRUCTEVAL:} towards holistic evaluation of instruction-tuned large language models.
\newblock \emph{CoRR}, abs/2306.04757, 2023.
\newblock \doi{10.48550/ARXIV.2306.04757}.
\newblock URL \url{https://doi.org/10.48550/arXiv.2306.04757}.

\bibitem[Cohen et~al.(2008)Cohen, Gimpel, and Smith]{NIPS2008_f11bec14}
Shay Cohen, Kevin Gimpel, and Noah~A Smith.
\newblock Logistic normal priors for unsupervised probabilistic grammar induction.
\newblock In D.~Koller, D.~Schuurmans, Y.~Bengio, and L.~Bottou (eds.), \emph{Advances in Neural Information Processing Systems}, volume~21. Curran Associates, Inc., 2008.
\newblock URL \url{https://proceedings.neurips.cc/paper_files/paper/2008/file/f11bec1411101c743f64df596773d0b2-Paper.pdf}.

\bibitem[Dyer et~al.(2016)Dyer, Kuncoro, Ballesteros, and Smith]{dyer-etal-2016-recurrent}
Chris Dyer, Adhiguna Kuncoro, Miguel Ballesteros, and Noah~A. Smith.
\newblock Recurrent neural network grammars.
\newblock In Kevin Knight, Ani Nenkova, and Owen Rambow (eds.), \emph{Proceedings of the 2016 Conference of the North {A}merican Chapter of the Association for Computational Linguistics: Human Language Technologies}, pp.\  199--209, San Diego, California, June 2016.
\newblock \doi{10.18653/v1/N16-1024}.
\newblock URL \url{N16-1024}.

\bibitem[Eisner(2016)]{eisner-2016-inside}
Jason Eisner.
\newblock Inside-outside and forward-backward algorithms are just backprop (tutorial paper).
\newblock In Kai-Wei Chang, Ming-Wei Chang, Alexander Rush, and Vivek Srikumar (eds.), \emph{Proceedings of the Workshop on Structured Prediction for {NLP}}, pp.\  1--17, Austin, TX, November 2016.
\newblock \doi{10.18653/v1/W16-5901}.
\newblock URL \url{W16-5901}.

\bibitem[Fisher(1915)]{Fisher_1915}
R.~A. Fisher.
\newblock Frequency distribution of the values of the correlation coefficient in samples from an indefinitely large population.
\newblock \emph{Biometrika}, 10\penalty0 (4):\penalty0 507, May 1915.
\newblock ISSN 00063444.
\newblock \doi{10.2307/2331838}.
\newblock URL \url{https://www.jstor.org/stable/2331838?origin=crossref}.

\bibitem[He et~al.(2020)He, Wang, and Zhang]{he-etal-2020-enhancing}
Qi~He, Han Wang, and Yue Zhang.
\newblock Enhancing generalization in natural language inference by syntax.
\newblock In Trevor Cohn, Yulan He, and Yang Liu (eds.), \emph{Findings of the Association for Computational Linguistics: EMNLP 2020}, pp.\  4973--4978, Online, November 2020. Association for Computational Linguistics.
\newblock \doi{10.18653/v1/2020.findings-emnlp.447}.
\newblock URL \url{https://aclanthology.org/2020.findings-emnlp.447}.

\bibitem[Heim \& Kratzer(1998)Heim and Kratzer]{Heim_Kratzer_1998}
Irene Heim and Angelika Kratzer.
\newblock \emph{Semantics in generative grammar}.
\newblock Blackwell textbooks in linguistics. Blackwell, Malden, MA, 1998.
\newblock ISBN 9780631197126.

\bibitem[Johnson et~al.(2007)Johnson, Griffiths, and Goldwater]{johnson-etal-2007-bayesian}
Mark Johnson, Thomas Griffiths, and Sharon Goldwater.
\newblock {B}ayesian inference for {PCFG}s via {M}arkov chain {M}onte {C}arlo.
\newblock In Candace Sidner, Tanja Schultz, Matthew Stone, and ChengXiang Zhai (eds.), \emph{Human Language Technologies 2007: The Conference of the North {A}merican Chapter of the Association for Computational Linguistics; Proceedings of the Main Conference}, pp.\  139--146, Rochester, New York, April 2007.
\newblock URL \url{N07-1018}.

\bibitem[Kim et~al.(2019{\natexlab{a}})Kim, Dyer, and Rush]{kim-etal-2019-compound}
Yoon Kim, Chris Dyer, and Alexander Rush.
\newblock Compound probabilistic context-free grammars for grammar induction.
\newblock In Anna Korhonen, David Traum, and Llu{\'\i}s M{\`a}rquez (eds.), \emph{Proceedings of the 57th Annual Meeting of the Association for Computational Linguistics}, pp.\  2369--2385, Florence, Italy, July 2019{\natexlab{a}}.
\newblock \doi{10.18653/v1/P19-1228}.
\newblock URL \url{P19-1228}.

\bibitem[Kim et~al.(2019{\natexlab{b}})Kim, Rush, Yu, Kuncoro, Dyer, and Melis]{kim-etal-2019-unsupervised}
Yoon Kim, Alexander Rush, Lei Yu, Adhiguna Kuncoro, Chris Dyer, and G{\'a}bor Melis.
\newblock Unsupervised recurrent neural network grammars.
\newblock In Jill Burstein, Christy Doran, and Thamar Solorio (eds.), \emph{Proceedings of the 2019 Conference of the North {A}merican Chapter of the Association for Computational Linguistics: Human Language Technologies, Volume 1 (Long and Short Papers)}, pp.\  1105--1117, Minneapolis, Minnesota, June 2019{\natexlab{b}}.
\newblock \doi{10.18653/v1/N19-1114}.
\newblock URL \url{N19-1114}.

\bibitem[Klein \& Manning(2002)Klein and Manning]{klein-manning-2002-generative}
Dan Klein and Christopher~D. Manning.
\newblock A generative constituent-context model for improved grammar induction.
\newblock In Pierre Isabelle, Eugene Charniak, and Dekang Lin (eds.), \emph{Proceedings of the 40th Annual Meeting of the Association for Computational Linguistics}, pp.\  128--135, Philadelphia, Pennsylvania, USA, July 2002.
\newblock \doi{10.3115/1073083.1073106}.
\newblock URL \url{P02-1017}.

\bibitem[Li et~al.(2007)Li, Fan, and Zhang]{Li_Fan_Zhang_2007}
Juanzi Li, Qi’na Fan, and Kuo Zhang.
\newblock Keyword extraction based on tf/idf for chinese news document.
\newblock \emph{Wuhan University Journal of Natural Sciences}, 12\penalty0 (5):\penalty0 917–921, September 2007.
\newblock ISSN 1007-1202, 1993-4998.
\newblock \doi{10.1007/s11859-007-0038-4}.
\newblock URL \url{http://link.springer.com/10.1007/s11859-007-0038-4}.

\bibitem[Liu et~al.(2023)Liu, Yang, Kim, and Tu]{liu-etal-2023-simple}
Wei Liu, Songlin Yang, Yoon Kim, and Kewei Tu.
\newblock Simple hardware-efficient {PCFG}s with independent left and right productions.
\newblock In Houda Bouamor, Juan Pino, and Kalika Bali (eds.), \emph{Findings of the Association for Computational Linguistics: EMNLP 2023}, pp.\  1662--1669, Singapore, December 2023. Association for Computational Linguistics.
\newblock \doi{10.18653/v1/2023.findings-emnlp.113}.
\newblock URL \url{https://aclanthology.org/2023.findings-emnlp.113}.

\bibitem[Marcus et~al.(1999)Marcus, Santorini, Marcinkiewicz, and Taylor]{Treebank-3}
Mitchell~P. Marcus, Beatrice Santorini, Mary~Ann Marcinkiewicz, and Ann Taylor.
\newblock Treebank-3, 1999.
\newblock URL \url{https://catalog.ldc.upenn.edu/LDC99T42}.

\bibitem[Minaee et~al.(2024)Minaee, Mikolov, Nikzad, Chenaghlu, Socher, Amatriain, and Gao]{Minaee2024LargeLM}
Shervin Minaee, Tom Mikolov, Narjes Nikzad, Meysam~Asgari Chenaghlu, Richard Socher, Xavier Amatriain, and Jianfeng Gao.
\newblock Large language models: A survey.
\newblock \emph{ArXiv}, abs/2402.06196, 2024.
\newblock URL \url{https://api.semanticscholar.org/CorpusID:267617032}.

\bibitem[Mishra \& Vishwakarma(2015)Mishra and Vishwakarma]{mishra2015analysis}
Apra Mishra and Santosh Vishwakarma.
\newblock Analysis of tf-idf model and its variant for document retrieval.
\newblock In \emph{2015 international conference on computational intelligence and communication networks (cicn)}, pp.\  772--776. IEEE, 2015.

\bibitem[Palmer et~al.(2005)Palmer, Chiou, Xue, and Lee]{CTB5.1}
Martha Palmer, Fu-Dong Chiou, Nianwen Xue, and Tsan-Kuang Lee.
\newblock Chinese treebank 5.0, January 2005.
\newblock URL \url{https://catalog.ldc.upenn.edu/LDC2005T01}.

\bibitem[Pollard \& Sag(1987)Pollard and Sag]{Pollard_Sag_1987}
Carl~Jesse Pollard and Ivan~A. Sag.
\newblock \emph{Information-based syntax and semantics}.
\newblock CSLI lecture notes. Center for the Study of Language and Information, Stanford, CA, 1987.
\newblock ISBN 9780937073230.

\bibitem[Schulman et~al.(2016)Schulman, Moritz, Levine, Jordan, and Abbeel]{DBLP:journals/corr/SchulmanMLJA15}
John Schulman, Philipp Moritz, Sergey Levine, Michael~I. Jordan, and Pieter Abbeel.
\newblock High-dimensional continuous control using generalized advantage estimation.
\newblock In Yoshua Bengio and Yann LeCun (eds.), \emph{4th International Conference on Learning Representations, {ICLR} 2016, San Juan, Puerto Rico, May 2-4, 2016, Conference Track Proceedings}, 2016.
\newblock URL \url{http://arxiv.org/abs/1506.02438}.

\bibitem[Seddah et~al.(2013)Seddah, Tsarfaty, K{\"u}bler, Candito, Choi, Farkas, Foster, Goenaga, Gojenola~Galletebeitia, Goldberg, Green, Habash, Kuhlmann, Maier, Nivre, Przepi{\'o}rkowski, Roth, Seeker, Versley, Vincze, Woli{\'n}ski, Wr{\'o}blewska, and Villemonte de~la Clergerie]{seddah-etal-2013-overview}
Djam{\'e} Seddah, Reut Tsarfaty, Sandra K{\"u}bler, Marie Candito, Jinho~D. Choi, Rich{\'a}rd Farkas, Jennifer Foster, Iakes Goenaga, Koldo Gojenola~Galletebeitia, Yoav Goldberg, Spence Green, Nizar Habash, Marco Kuhlmann, Wolfgang Maier, Joakim Nivre, Adam Przepi{\'o}rkowski, Ryan Roth, Wolfgang Seeker, Yannick Versley, Veronika Vincze, Marcin Woli{\'n}ski, Alina Wr{\'o}blewska, and Eric Villemonte de~la Clergerie.
\newblock Overview of the {SPMRL} 2013 shared task: A cross-framework evaluation of parsing morphologically rich languages.
\newblock In Yoav Goldberg, Yuval Marton, Ines Rehbein, and Yannick Versley (eds.), \emph{Proceedings of the Fourth Workshop on Statistical Parsing of Morphologically-Rich Languages}, pp.\  146--182, Seattle, Washington, USA, October 2013.
\newblock URL \url{W13-4917}.

\bibitem[Shayegh et~al.(2024)Shayegh, Cao, Zhu, Cheung, and Mou]{DBLP:conf/iclr/ShayeghC0CM24}
Behzad Shayegh, Yanshuai Cao, Xiaodan Zhu, Jackie C.~K. Cheung, and Lili Mou.
\newblock Ensemble distillation for unsupervised constituency parsing.
\newblock In \emph{The Twelfth International Conference on Learning Representations, {ICLR} 2024, Vienna, Austria, May 7-11, 2024}. OpenReview.net, 2024.
\newblock URL \url{https://openreview.net/forum?id=RR8y0WKrFv}.

\bibitem[Shen et~al.(2017)Shen, Lin, Huang, and Courville]{Shen2017NeuralLM}
Yikang Shen, Zhouhan Lin, Chin-Wei Huang, and Aaron~C. Courville.
\newblock Neural language modeling by jointly learning syntax and lexicon.
\newblock \emph{ArXiv}, abs/1711.02013, 2017.
\newblock URL \url{https://api.semanticscholar.org/CorpusID:3347806}.

\bibitem[Silver \& Dunlap(1987)Silver and Dunlap]{Silver_Dunlap_1987}
N.~Clayton Silver and William~P. Dunlap.
\newblock Averaging correlation coefficients: Should fisher’s z transformation be used?
\newblock \emph{Journal of Applied Psychology}, 72\penalty0 (1):\penalty0 146–148, February 1987.
\newblock ISSN 1939-1854, 0021-9010.
\newblock \doi{10.1037/0021-9010.72.1.146}.
\newblock URL \url{http://doi.apa.org/getdoi.cfm?doi=10.1037/0021-9010.72.1.146}.

\bibitem[Sparck~Jones(1972)]{Sparck_Jones_1972}
Karen Sparck~Jones.
\newblock A statistical interpretation of term specificity and its application in retrieval.
\newblock \emph{Journal of Documentation}, 28\penalty0 (1):\penalty0 11–21, January 1972.
\newblock ISSN 0022-0418.
\newblock \doi{10.1108/eb026526}.
\newblock URL \url{https://www.emerald.com/insight/content/doi/10.1108/eb026526/full/html}.

\bibitem[Spearman(1904)]{Spearman_1904}
C.~Spearman.
\newblock The proof and measurement of association between two things.
\newblock \emph{The American Journal of Psychology}, 15\penalty0 (1):\penalty0 72, January 1904.
\newblock ISSN 00029556.
\newblock \doi{10.2307/1412159}.
\newblock URL \url{https://www.jstor.org/stable/1412159?origin=crossref}.

\bibitem[Spitkovsky et~al.(2010)Spitkovsky, Alshawi, Jurafsky, and Manning]{spitkovsky-etal-2010-viterbi}
Valentin~I. Spitkovsky, Hiyan Alshawi, Daniel Jurafsky, and Christopher~D. Manning.
\newblock {V}iterbi training improves unsupervised dependency parsing.
\newblock In Mirella Lapata and Anoop Sarkar (eds.), \emph{Proceedings of the Fourteenth Conference on Computational Natural Language Learning}, pp.\  9--17, Uppsala, Sweden, July 2010.
\newblock URL \url{W10-2902}.

\bibitem[Steedman(2000)]{Steedman_2000}
Mark Steedman.
\newblock \emph{The syntactic process}.
\newblock Language, speech, and communication. MIT Press, Cambridge, Mass, 2000.
\newblock ISBN 9780262194204.

\bibitem[Stern et~al.(2017)Stern, Andreas, and Klein]{stern-etal-2017-minimal}
Mitchell Stern, Jacob Andreas, and Dan Klein.
\newblock A minimal span-based neural constituency parser.
\newblock In Regina Barzilay and Min-Yen Kan (eds.), \emph{Proceedings of the 55th Annual Meeting of the Association for Computational Linguistics (Volume 1: Long Papers)}, pp.\  818--827, Vancouver, Canada, July 2017.
\newblock \doi{10.18653/v1/P17-1076}.
\newblock URL \url{P17-1076}.

\bibitem[Williams(1992)]{DBLP:journals/ml/Williams92}
Ronald~J. Williams.
\newblock Simple statistical gradient-following algorithms for connectionist reinforcement learning.
\newblock \emph{Mach. Learn.}, 8:\penalty0 229--256, 1992.
\newblock \doi{10.1007/BF00992696}.
\newblock URL \url{https://doi.org/10.1007/BF00992696}.

\bibitem[Xie \& Xing(2017)Xie and Xing]{xie-xing-2017-constituent}
Pengtao Xie and Eric Xing.
\newblock A constituent-centric neural architecture for reading comprehension.
\newblock In Regina Barzilay and Min-Yen Kan (eds.), \emph{Proceedings of the 55th Annual Meeting of the Association for Computational Linguistics (Volume 1: Long Papers)}, pp.\  1405--1414, Vancouver, Canada, July 2017.
\newblock \doi{10.18653/v1/P17-1129}.
\newblock URL \url{P17-1129}.

\bibitem[Xu et~al.(2021)Xu, Drozdov, Lee, O{'}Gorman, Rongali, Finkbeiner, Suresh, Iyyer, and McCallum]{xu-etal-2021-improved}
Zhiyang Xu, Andrew Drozdov, Jay~Yoon Lee, Tim O{'}Gorman, Subendhu Rongali, Dylan Finkbeiner, Shilpa Suresh, Mohit Iyyer, and Andrew McCallum.
\newblock Improved latent tree induction with distant supervision via span constraints.
\newblock In Marie-Francine Moens, Xuanjing Huang, Lucia Specia, and Scott Wen-tau Yih (eds.), \emph{Proceedings of the 2021 Conference on Empirical Methods in Natural Language Processing}, pp.\  4818--4831, Online and Punta Cana, Dominican Republic, November 2021. Association for Computational Linguistics.
\newblock \doi{10.18653/v1/2021.emnlp-main.395}.
\newblock URL \url{https://aclanthology.org/2021.emnlp-main.395}.

\bibitem[Yang et~al.(2021{\natexlab{a}})Yang, Zhao, and Tu]{yang-etal-2021-neural}
Songlin Yang, Yanpeng Zhao, and Kewei Tu.
\newblock Neural bi-lexicalized {PCFG} induction.
\newblock In Chengqing Zong, Fei Xia, Wenjie Li, and Roberto Navigli (eds.), \emph{Proceedings of the 59th Annual Meeting of the Association for Computational Linguistics and the 11th International Joint Conference on Natural Language Processing (Volume 1: Long Papers)}, pp.\  2688--2699, Online, August 2021{\natexlab{a}}. Association for Computational Linguistics.
\newblock \doi{10.18653/v1/2021.acl-long.209}.
\newblock URL \url{https://aclanthology.org/2021.acl-long.209}.

\bibitem[Yang et~al.(2021{\natexlab{b}})Yang, Zhao, and Tu]{yang-etal-2021-pcfgs}
Songlin Yang, Yanpeng Zhao, and Kewei Tu.
\newblock {PCFG}s can do better: Inducing probabilistic context-free grammars with many symbols.
\newblock In Kristina Toutanova, Anna Rumshisky, Luke Zettlemoyer, Dilek Hakkani-Tur, Iz~Beltagy, Steven Bethard, Ryan Cotterell, Tanmoy Chakraborty, and Yichao Zhou (eds.), \emph{Proceedings of the 2021 Conference of the North American Chapter of the Association for Computational Linguistics: Human Language Technologies}, pp.\  1487--1498, Online, June 2021{\natexlab{b}}. Association for Computational Linguistics.
\newblock \doi{10.18653/v1/2021.naacl-main.117}.
\newblock URL \url{https://aclanthology.org/2021.naacl-main.117}.

\bibitem[Zhang et~al.(2021)Zhang, Song, Jin, Xu, Yu, and Luo]{zhang-etal-2021-video}
Songyang Zhang, Linfeng Song, Lifeng Jin, Kun Xu, Dong Yu, and Jiebo Luo.
\newblock Video-aided unsupervised grammar induction.
\newblock In Kristina Toutanova, Anna Rumshisky, Luke Zettlemoyer, Dilek Hakkani-Tur, Iz~Beltagy, Steven Bethard, Ryan Cotterell, Tanmoy Chakraborty, and Yichao Zhou (eds.), \emph{Proceedings of the 2021 Conference of the North American Chapter of the Association for Computational Linguistics: Human Language Technologies}, pp.\  1513--1524, Online, June 2021. Association for Computational Linguistics.
\newblock \doi{10.18653/v1/2021.naacl-main.119}.
\newblock URL \url{https://aclanthology.org/2021.naacl-main.119}.

\bibitem[Zhao \& Titov(2020)Zhao and Titov]{zhao-titov-2020-visually}
Yanpeng Zhao and Ivan Titov.
\newblock Visually grounded compound {PCFG}s.
\newblock In Bonnie Webber, Trevor Cohn, Yulan He, and Yang Liu (eds.), \emph{Proceedings of the 2020 Conference on Empirical Methods in Natural Language Processing (EMNLP)}, pp.\  4369--4379, Online, November 2020. Association for Computational Linguistics.
\newblock \doi{10.18653/v1/2020.emnlp-main.354}.
\newblock URL \url{https://aclanthology.org/2020.emnlp-main.354}.

\bibitem[Ziebart et~al.(2008)Ziebart, Maas, Bagnell, and Dey]{Ziebart2008MaximumEI}
Brian~D. Ziebart, Andrew~L. Maas, J.~Andrew Bagnell, and Anind~K. Dey.
\newblock Maximum entropy inverse reinforcement learning.
\newblock In \emph{AAAI Conference on Artificial Intelligence}, 2008.
\newblock URL \url{https://api.semanticscholar.org/CorpusID:336219}.

\end{thebibliography}
\bibliographystyle{src/iclr2025_conference}

\newpage
\appendix
\section{Appendix}
\subsection{Advanced SemInfo Maximization}
\label{appendix:asm}
In Section~\ref{sec:treecrf}, we presented a SemInfo maximization method that performs mean-field optimization through a TreeCRF model.
This method parameterizes a TreeCRF model using the span-posterior probability and maximizes the expected SemInfo value of the TreeCRF distribution.
While our study demonstrated significant accuracy improvement by the TreeCRF-based SemInfo maximization training, it raises a new question: What is the advantage of optimizing the PCFG parameters through the TreeCRF model compared to optimizing those parameters directly?
In the experiment presented in this section, we found no significant difference between the TreeCRF-based and PCFG-based optimizations.
It shows that the demonstrated accuracy improvement does not depend on particular optimization methods, highlighting the contribution of the SemInfo maximization objective to accurate unsupervised parsing.
In addition, we found that the TreeCRF-based optimization is more time and space-efficient than the PCFG-based optimization, which makes the TreeCRF-based optimization preferable.

As shown in Table~\ref{tbl:lookup}, we compare three SemInfo maximization strategies combined with two optimization methods (TreeCRF-based and PCFG-based methods). 
We include an LL-trained parser and a supervised parser as baselines. 
The two baselines serve as the lower and upper bounds for the comparison, respectively.

\begin{table}[t]
    \centering
    % \adjustbox{width=\linewidth}{
    \begin{tabular}{l|ccc}
    \hline
        Strategy & RL Baseline & Sampling distribution & Learning Strategy \\ \hline
        Action-V & V-function & Action distribution & SemInfo maximization  \\ 
        Posterior-V & V-function & Tree posterior distribution & SemInfo maximization  \\ 
        Posterior-Avg & Sample average & Tree posterior distribution & SemInfo maximization\\ \hline
        Supervised & - & - &  Supervised learning \\
        % PCFG& - &&supervised& ref1 \\\hline
        LL & - & - &  Likelihood maximization  \\\hline
    \end{tabular}
    \caption{Lookup table for optimization strategies and detailed descriptions}
    \label{tbl:lookup}
\end{table}

% \subsubsection{Sampling from $P^{CRF}(t|x)$ and $P^{PCFG}(t|x)$}
\subsubsection{Tree Posterior and Sampling Distributions}
Both the TreeCRF-based and PCFG-based optimizations aim to maximize the expected SemInfo with on-policy Reinforcement Learning (RL), but they operate on two policy distributions: the TreeCRF-based posterior distribution $P^{CRF}(t|x)$ (Equation~\ref{eq:TreeCRF-potential}) and the PCFG-based posterior distribution $P^{PCFG}(t|x)$ (Equation~\ref{eq:pcfg-posterior}).
Sampling from both distributions involves multiple span-splitting steps (Algorithm~\ref{alg:CRF-sampler} for the TreeCRF model and Algorithm~\ref{alg:PCFG-sampler} for the PCFG model).
The sampler starts with the sentence-level span ($(1, n)$ for the TreeCRF model; $(S, 1, n)$ for the PCFG model) and recursively makes splitting decisions ($(i, j)\to (i, k) (k, j)$ for the TreeCRF model; $(A, i, j)\to (B, i, k) (C, k, j)$ for the PCFG model).
The sampler repeats this span-splitting process until it reaches single-word spans ($j=i+1$).

\begin{small}
\begin{equation}
    P^{PCFG}(t|x)= \frac{P(x, t)}{\sum_t P(x, t)} \label{eq:pcfg-posterior}
\end{equation}
\end{small}

\begin{algorithm}[t]
\footnotesize
\caption{TreeCRF Sampler}
\label{alg:CRF-sampler}
\begin{algorithmic}[1]
\STATE \textbf{function} CRF-Sampler($i, j, x$)
    \IF{$j = i + 1$} 
        \STATE \textbf{Return} leaf node $(i, j)$
    \ELSE
        \STATE Sample split index $k \sim \pi^{CRF}(k \mid (i,j))$ following Equation~\ref{eq:CRF-sampling-policy} \cite{johnson-etal-2007-bayesian}
        \STATE $T_{\text{left}} \gets$ CRF-Sampler($i, k, x$)
        \STATE $T_{\text{right}} \gets$ CRF-Sampler($k, j, x$)
        \STATE \textbf{Return} node $(i,j)$ with children $T_{\text{left}}$ and $T_{\text{right}}$
    \ENDIF
\STATE \textbf{end function}
\end{algorithmic}
\end{algorithm}

\begin{small}
\begin{equation}
    \pi^{CRF}(k | (i, j)) = \frac{\exp(\sum_{s\leq (i, k)}\log P(s \text{ is a const.}|x) + \sum_{s\leq (k, j)}\log P(s \text{ is a const.}|x))}{\sum_k\exp(\sum_{s\leq (i, k)}\log P(s \text{ is a const.}|x) + \sum_{s\leq (k, j)}\log P(s \text{ is a const.}|x))}
    \label{eq:CRF-sampling-policy}
\end{equation}    
\end{small}

\begin{algorithm}[t]
\footnotesize
\caption{PCFG Sampler}
\label{alg:PCFG-sampler}
\begin{algorithmic}[1]
\STATE \textbf{function} PCFG-Sampler($A, i, j, x$)
    \IF{$j = i + 1$} 
        \STATE \textbf{Return} leaf node $(A, i, j)$
    \ELSE
        \STATE Sample split index $B, C, k \sim \pi^{PCFG}(B, C, k \mid (A, i,j))$ following Equation~\ref{eq:PCFG-sampling-policy}
        \STATE $T_{\text{left}} \gets$ PCFG-Sampler($B, i, k, x$)
        \STATE $T_{\text{right}} \gets$ PCFG-Sampler($C, k, j, x$)
        \STATE \textbf{Return} node $(A, i,j)$ with children $T_{\text{left}}$ and $T_{\text{right}}$
    \ENDIF
\STATE \textbf{end function}
\end{algorithmic}
\end{algorithm}

\begin{small}
\begin{equation}
    \pi^{PCFG}(B, C, k | (A, i, j)) = \frac{P(A\to B C)\beta(B, i, k)\beta(C, k, j)}{\beta(A, i, j)}
    \label{eq:PCFG-sampling-policy}
\end{equation}
\end{small}

\subsubsection{Three SemInfo Maximization Strategies}
In this subsection, we introduce two RL optimization methods: posterior and action optimizations.
We further combine the optimization methods with two RL baseline estimations: average baseline and $V$-function baselines. 
Since the average baseline cannot be applied to the action optimization, the combination of the optimizations and baselines gives three optimization strategies (Table~\ref{tbl:lookup}).

The posterior optimization is similar to the method explained in the main text: (1) sampling tree from either $P^{CRF}(t|x)$ or $P^{PCFG}(t|x)$; and (2) perform policy gradient optimization in accordance with Equation~\ref{eq:loss}.
In contrast, the action optimization considers the tree sampling process as an RL trajectory and applies the SemInfo maximization through $\pi^{CRF}$ or $\pi^{PCFG}$.
The two optimizations differ in how the RL agent is defined. 
The posterior optimization defines the RL agent as a one-step agent and seeks to maximize the SemInfo values for the parser-predicted trees.
The action optimization defines the RL agent as a span-splitting agent and seeks to maximize the SemInfo for the tree resulting from the span-splitting decision.

\subsubsection{$V$-function Computation}

In the PCFG setting, the $V$-function can be computed precisely and efficiently using dynamic programming.
The $V$-function estimates the expected return of visiting a state $s$ ($s=(i, j)$ for the TreeCRF model and $s=(A, i, j)$ for the PCFG model) using a policy $\pi$ (Equation~\ref{eq:v-function}).
It enables the estimation of the advantage function $A(s, a)$, which evaluates how effective $a$ is in maximizing the SemInfo value of the sampled tree (Equation~\ref{eq:advantage-function}).
Algorithm~\ref{algo:TreeCRF-v-computation} and Algorithm~\ref{algo:PCFG-v-computation} details the $V$-function computation for both the TreeCRF and PCFG-based optimizations.
The two algorithms share the same backbone but differ in the span-splitting agent $\pi$.

\begin{small}
\begin{equation}
    V(s) = \mathop{\E}_{(s_0, a_0, s_1, a_1,\dots)\sim\pi}\left[\sum_{(s_i, a_i)}r(s_i, a_i)\right] \label{eq:v-function}
\end{equation}
\begin{equation}
    A(s, a) = r(s, a) + V(s^\prime) - V(s)\label{eq:advantage-function}
\end{equation}
\end{small}

% (Algorithm~\ref{} for the TreeCRF setting and Algorithm~\ref{} for the PCFG setting).

\begin{algorithm}[t]
\footnotesize
\caption{TreeCRF-based V-Function Computation}
\label{algo:TreeCRF-v-computation}
\begin{algorithmic}[1]
\REQUIRE Subtree-selection agent $\pi^{CRF}(i, j)$ governing all possible split decisions of span $(i, j)\to (i, k) (k, j)$. 
\REQUIRE SemInfo function $r(i, j) = I(x_{i:j}, Sem(x))$.
\ENSURE $V(i, j)$ for all spans. 
\STATE Initialize $V(i, j) = 0$ for all spans $(i, j)$.
\STATE $w \gets 2$
\REPEAT
    \FOR{each span $s$ of length $w$}
        \STATE $V(s) \gets \begin{cases}
            0 &  w=1 \\
            r(i, j) &  w=2\\
            \E_{\pi^{CRF}(k|s)}\left[ (V(i, k)+V(k, j))\right] + r(i, j) &  w>2
        \end{cases}$
    \ENDFOR
    \STATE $w \gets w + 1$
\UNTIL{$w = n$}
\RETURN $V$
\end{algorithmic}
\end{algorithm}

\begin{algorithm}[t]
\footnotesize
\caption{PCFG-based V-Function Computation}
\label{algo:PCFG-v-computation}
\begin{algorithmic}[1]
\REQUIRE Subtree-selection agent $\pi^{PCFG}(A, i, j)$ governing all possible split decisions of span $(A, i, j)\to (B, i, k) (C, k, j)$. 
\REQUIRE SemInfo function $r(A, i, j) = I(x_{i:j}, Sem(x))$.
\ENSURE $V(A, i, j)$. 
\STATE Initialize $V(A, i, j) = 0$ for all $(A, i, j)$.
\STATE $w \gets 2$
\REPEAT
    \FOR{each span $s$ of length $w$}
        \STATE $V(s) \gets \begin{cases}
            0 &  w=1 \\
            r(A, i, j) &  w=2\\
            \mathop{\E}_{\pi^{PCFG}(B, C, k|A, i, j)}\left[ (V(B, i, k)+V(C, k, j))\right] + r(A, i, j) &  w>2
        \end{cases}$
    \ENDFOR
    \STATE $w \gets w + 1$
\UNTIL{$w = n$}
\RETURN $V$
\end{algorithmic}
\end{algorithm}

\subsubsection{TreeCRF-based Optimizations}
Equation~\ref{eq:tcrf-posterior-avg} (restatement of Equation~\ref{eq:loss}), \ref{eq:tcrf-posterior-v}, and \ref{eq:tcrf-stepwise} details the training objective for the TreeCRF-based Posterior-Avg, Posterior-V, and Action-V strategies, respectively. 
Comparing Equation~\ref{eq:tcrf-posterior-avg} and Equation~\ref{eq:tcrf-posterior-v}, the Posterior-V improves over the Posterior-Avg by substituting the sample-average baseline with the $V$-function baseline. 
Our preliminary experiments indicate that the substitution results in faster convergence and slightly higher parsing accuracy. 
The Action-V strategy directly optimizes $\pi^{CRF}$ using the advantage function $A$, which we further augment with the Generalized Advantage Estimation \cite{DBLP:journals/corr/SchulmanMLJA15}.  

\begin{small}
    \begin{equation}
        \begin{aligned}
            \gJ(\gD) =  \mathop{\E}_{x\sim \mathcal{D}}[\log Z(x) + \mathop{\E}_{t\sim P^{CRF}(t|x)}[&\log P^{CRF}(t|x) (I(t, Sem(x)) -\\
    &\mathop{\E}_{t\sim P^{CRF}(t|x)}I(t, Sem(x)) + \beta H(P(t|x)))]]\label{eq:tcrf-posterior-avg}
        \end{aligned}
    \end{equation}
    % \vfill
    \vspace{0.5cm}
    \begin{equation}
        \begin{aligned}
            \gJ(\gD) =  \mathop{\E}_{x\sim \mathcal{D}}[\log Z(x) + \mathop{\E}_{t\sim P^{CRF}(t|x)}[&\log P^{CRF}(t|x) (I(t, Sem(x)) -\\&V_x(1, n) + \beta H(P^{CRF}(t|x)))]]\label{eq:tcrf-posterior-v}
        \end{aligned}
    \end{equation}
    \vspace{0.5cm}
    \begin{equation}
        \begin{aligned}
            \gJ(\gD) =  \mathop{\E}_{x\sim \mathcal{D}}[\log Z(x) + \mathop{\E}_{((1, n), k, \dots)\sim \pi^{CRF}}[&\sum_{(i, j), k}\log \pi^{CRF}(k|i, j) (A(i, j, k) \\&+ \beta H(\pi^{CRF}(k|i, j)))]]\label{eq:tcrf-stepwise}
        \end{aligned}
    \end{equation}
\end{small}

\subsubsection{PCFG-based Optimizations}
Equation~\ref{eq:pcfg-posterior-avg}, \ref{eq:pcfg-posterior-v}, and \ref{eq:pcfg-stepwise} details the training objective for the PCFG-based Posterior-Avg, Posterior-V, and Action-V strategies, respectively. 
These strategies are defined similarly to the TreeCRF-based strategies but replace the TreeCRF-based distributions with the PCFG-based distributions. 

\begin{small}
    \begin{equation}
        \begin{aligned}
            \gJ(\gD) =  \mathop{\E}_{x\sim \mathcal{D}}[\log Z(x) + \mathop{\E}_{t\sim P^{PCFG}(t|x)}[&\log P^{PCFG}(t|x) (I(t, Sem(x)) -\\
    &\mathop{\E}_{t\sim P^{PCFG}(t|x)}I(t, Sem(x)) )]]\label{eq:pcfg-posterior-avg}
        \end{aligned}
    \end{equation}
    \vspace{0.5cm}
    % \vfill
    \begin{equation}
        \begin{aligned}
            \gJ(\gD) =  \mathop{\E}_{x\sim \mathcal{D}}[\log Z(x) + \mathop{\E}_{t\sim P^{PCFG}(t|x)}[&\log P^{PCFG}(t|x) (I(t, Sem(x)) - V_x(S, 1, n))]]\label{eq:pcfg-posterior-v}
        \end{aligned}
    \end{equation}
    \vspace{0.5cm}
    % \vfill
    \begin{equation}
        \begin{aligned}
            \gJ(\gD) =  \mathop{\E}_{x\sim \mathcal{D}}[&\log Z(x) + \mathop{\E}_{((A, 1, n), (B, C, k), \dots)\sim \pi^{PCFG}}[\\&\sum_{(A, i, j), (B, C, k)}\log \pi^{PCFG}(B, C, k|A, i, j) A(A, i, j, B, C, k)]]\label{eq:pcfg-stepwise}
        \end{aligned}
    \end{equation}
\end{small}

\subsubsection{Result}

\begin{table}[t]
% \vspace{-0.5cm}
    \centering
    \small
    % \begin{adjustbox}{width=\textwidth}
    \begin{tabular}{l|cc|cc}
    \hline
         & \multicolumn{2}{|c|}{NPCFG}  & \multicolumn{2}{c}{SNPCFG}     \\ \cline{2-5}
        ~ & TreeCRF & PCFG & TreeCRF & PCFG \\ \hline
        Action-V & 65.16±1.15 & 62.66±1.76 & 67.21±0.33 & 66.04±0.38 \\ 
        Posterior-V & 66.82±0.32 & 66.37±1.71 & 67.87±0.54 & 66.32±1.39 \\ 
        Posterior-Avg & 65.55±0.75 & 65.64±1.34 & 66.77±0.14 & 66.85±0.32 \\ \hline
        Supervised & 69.05±0.55 & 73.54±0.11 & 71.83±0.21 & 74.78±0.23 \\ 
        LL & \multicolumn{2}{|c|}{53.34±0.59}& \multicolumn{2}{c}{57.84±2.61}  \\\hline
    \end{tabular}
    % \end{adjustbox}
    \caption{\csfo\space scores of two PCFG variants trained combined with three SemInfo maximization strategies. We retrained the TreeCRF-based Posterior-Avg model and the LL model in this experiment. }
    \label{tbl:val-optim-strategy}

% \vspace{-0.5cm}
    
\end{table}

Table~\ref{tbl:val-optim-strategy} evaluates the TreeCRF and PCFG-based optimization methods using two PCFG variants. 
Both methods yield parsers of similar performance. 
The TreeCRF-based method yields higher mean parsing accuracy than the PCFG-based method, yet the difference is within the margin of error. 
All combinations yield parsers with 65~67\csfo\nspace scores, except for the NPCFG+Action-V+PCFG-based optimization combination.
This combination results in parsers with 62.66 mean \csfo\nspace score, massively underperforming other combinations.
The underperformance might be related to the low model capacity of the PCFG model, as we did not observe similar performance degradation in the SNPCFG model and other high-capacity models tested in our preliminary experiment.
Overall, the comparison disentangled the high accuracy of the SemInfo-trained PCFGs from specific optimization algorithms. 
It highlights the contribution of the SemInfo maximization objective to accurate unsupervised parsing.

In comparison with the supervised baseline, all combinations yield parsers with accuracy within 8 \csfo\nspace scores from the supervised baseline.  
The small gap, on the one hand, showcases the strong performance of the SemInfo-trained PCFG parsers. 
On the other hand, it indicates that the PCFG model might limit further development of semantic-aware unsupervised parsers. 
A more expressive parsing model (e.g., the Recurrent Neural Network Model \cite{dyer-etal-2016-recurrent}) might be necessary in future studies. 

While the TreeCRF-based and PCFG-based methods yield parsers of similar accuracies, we found the TreeCRF-based optimization more time and space-efficient than the PCFG-based optimization. 
The TreeCRF-based optimization trains NPCFG parsers at 4x the speed and uses only $\frac{1}{6}$ the memory. 
It also trains SNPCFG parsers at 8x the speed and uses $\frac{1}{8}$ the memory. 
The improved training efficiency enables the further scaling of the PCFG model.

\subsection{Computing Span-posterior probability via back-propagation}
\label{appendix:span-posterior-by-bp}
This section explains how the span-posterior probability $P(s\text{ is a constituent}|x)$ is computed using back-propagation.
\begin{equation}
    P(s\text{ is a constituent}|x) = \sum_{A\in NT} \frac{\partial \log Z(x)}{\partial \log \beta(s, A)}\label{eq:ioasbp-target}
\end{equation}
\begin{proof}
Firstly, we define the span-posterior probability as Equation~\ref{eq:spanposterior-def}. Here $s$ is a substring of $x$, spanning from the $i$-th word to the $j$-th word (i.e., $s:=(x_i,...,x_j)$). Intuitively, $s$ is a constituent if there exists a non-terminal $A$ that expands into $s$.
\begin{equation}
    P(s\text{ is a constituent}|x) = \frac{\sum_{A\in NT}P(S\rightarrow x \wedge A\rightarrow s_{i,j})}{P(x)}\label{eq:spanposterior-def}
\end{equation}
We split $P(S\rightarrow x \wedge A\rightarrow s_{i,j})$ into two parts in Equation~\ref{eq:inside-outside-split}: $P(S\rightarrow x_1, ..., x_{i-1}, A, x_{j+1},...)$, the probability of generating words \emph{outside} $s$, and $P(A\rightarrow s)$, the probability generating words \emph{inside} $s$. 
The outside probability can be computed using back-propagation \citep{eisner-2016-inside}.
The inside probability is already computed by the $\beta$ table.
Exploiting algebraic transformations shown in Equation~\ref{eq:ioasbp-alge}, we can derive the formula shown in Equation~\ref{eq:ioasbp-target}.
\begin{align}
    P(s\text{ is a constituent}|x) &= \frac{1}{Z(x)} \sum_{A\in NT}P(S\rightarrow x_1, ..., x_{i-1}, A, x_{j+1},...)P(A\rightarrow s)\label{eq:inside-outside-split}\\
    &=\frac{1}{Z(x)} \sum_{A\in NT}\frac{\partial Z(x)}{\partial \beta(s, A)}\beta(s, A)\\
    &=\frac{1}{Z(x)} \sum_{A\in NT}Z(x)\frac{\partial \log Z(x)}{\partial \log \beta(s, A)} \frac{1}{\beta(s, A)}\beta(s, A)\label{eq:ioasbp-alge}\\
    &=\sum_{A\in NT}\frac{\partial\log Z(X)}{\partial\log \beta(s, A)}\label{eq:ioasbp}
\end{align}
\end{proof}

\subsection{Sentence-level Correlation in different training stages}
\label{appendix:sent-corr-stages}
\begin{figure}[t]
    \centering
    \includegraphics[width=1\linewidth]{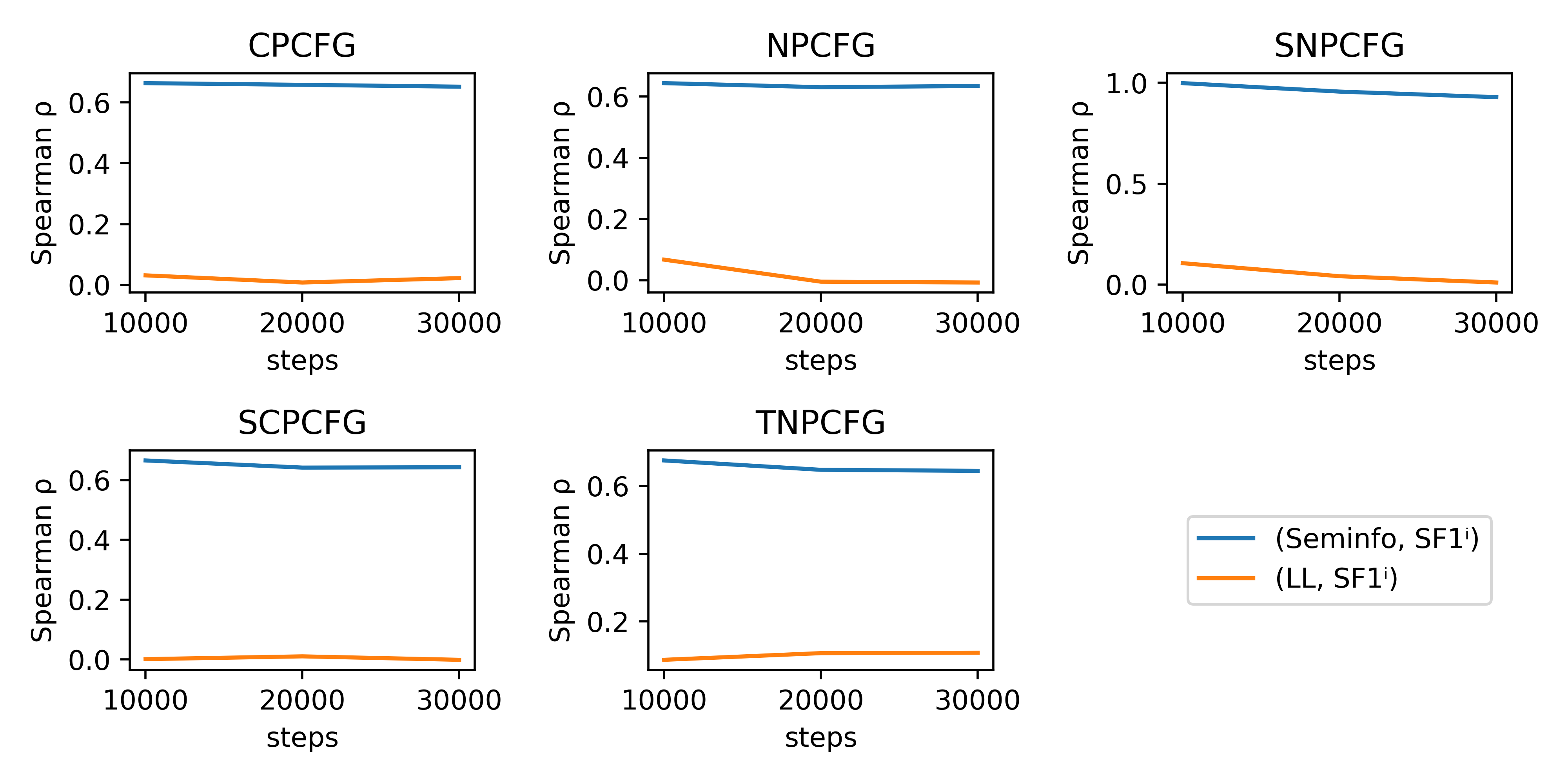}
    \caption{Sentence-level Spearman correlations for models trained for 10k steps, 20k steps, and 30k steps.}
    \label{fig:corr-sent-stages}
\end{figure}
\begin{figure}
    \centering
    \includegraphics[width=\linewidth]{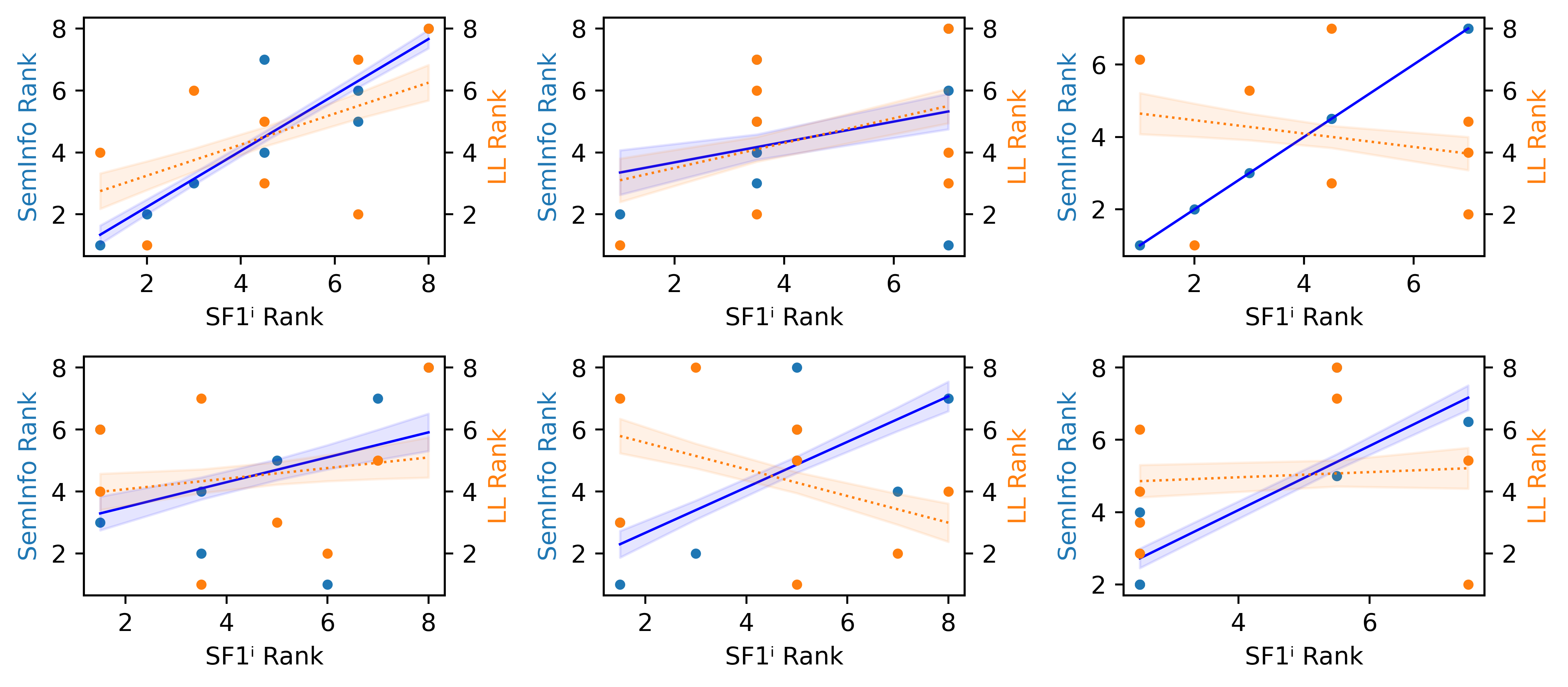}
    \caption{Sentence-level correlation on six random sentences.}
    \label{fig:sent_level-corr-6figs}
\end{figure}
In Table~\ref{tbl:corr}, we showed a strong sentence-level correlation between SemInfo and \isfo\nspace but a weak correlation between LL and \isfo.
Nevertheless, it remains unclear whether the correlation gap is related to the number of training steps 
% the question of whether the correlation gap is related to the number of training steps remains. 
Figure~\ref{fig:corr-sent-stages} excludes the number of training steps as a factor in the correlation gap. 
In this experiment, we calculate the correlation coefficient for models trained for 10k steps, 20k steps, and 30k steps. 
We can observe that, for all PCFG variants, the correlation coefficients for (SemInfo, \isfo) are consistently over 0.6, while the coefficients for (LL, \isfo) are consistently below 0.1. 
This result underscores our conclusion that SemInfo can serve as an accurate estimate of parsing accuracy.

\subsection{More detailed analysis for corpus-level correlation}
\label{appendix:corpus-corr}
\begin{figure}[t]
    \centering
    \includegraphics[width=0.9\linewidth]{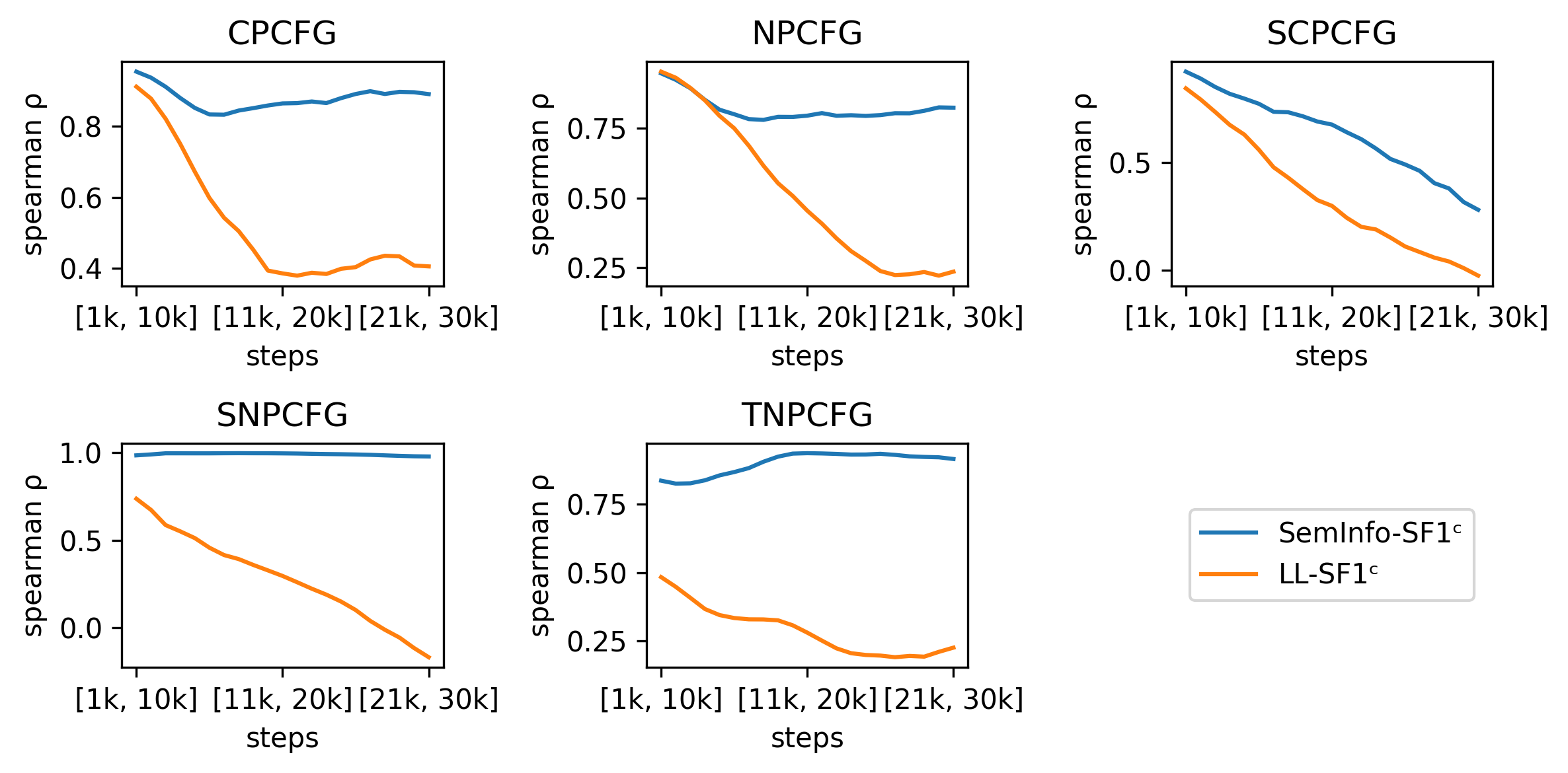}
    \caption{Corpus-level Spearman correlation in different training stages.}
    \label{fig:corr-corpus-training_stages}
\end{figure}
\begin{figure}
    \centering
    \includegraphics[width=1\linewidth]{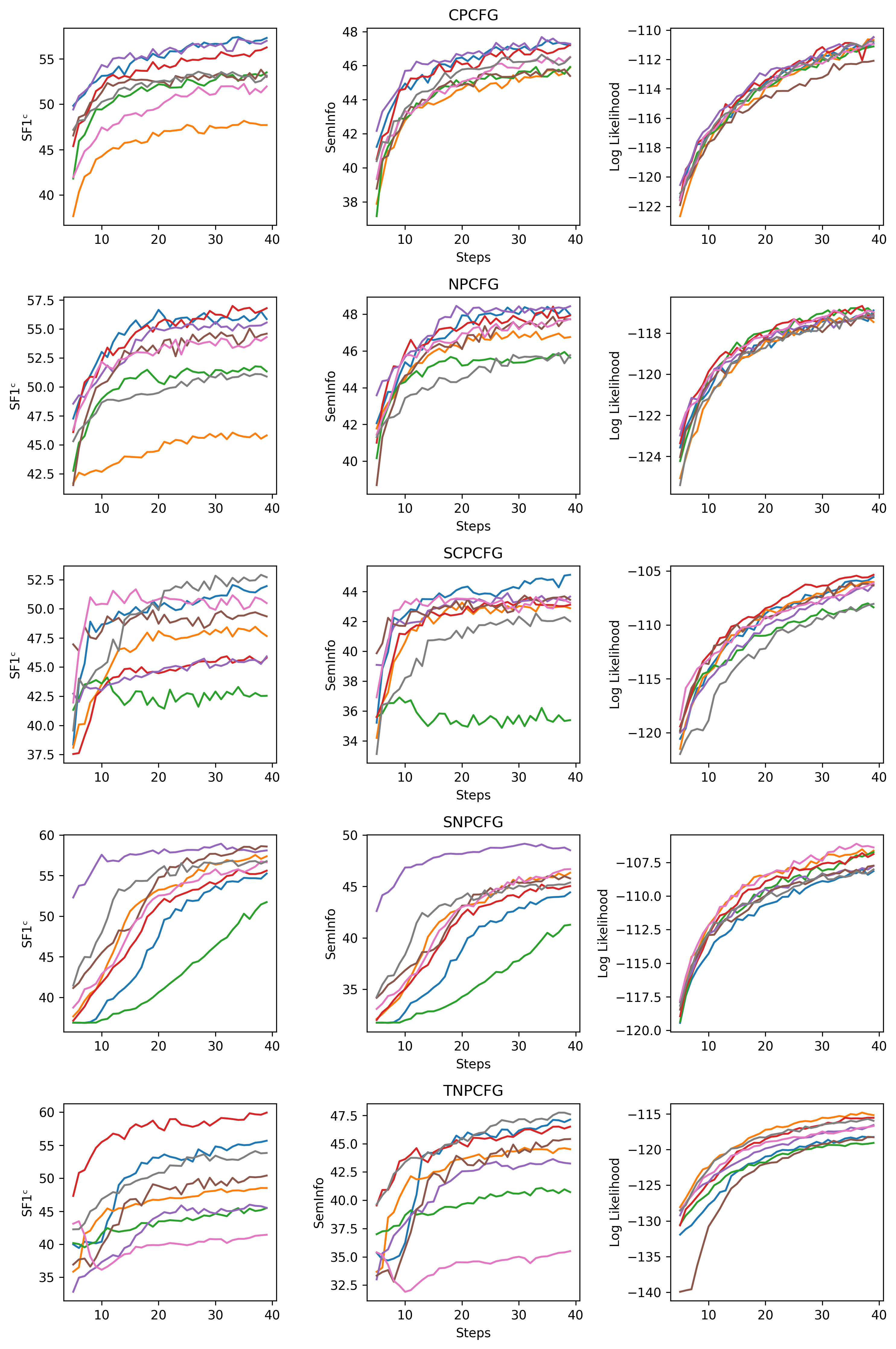}
    \caption{Training curves of SemInfo, LL, and \csfo. Each line represents the curve for a single PCFG model.}
    \label{fig:training-dynamics}
\end{figure}

Figure~\ref{fig:corr-corpus-training_stages} shows the corpus-level correlation in different training stages for all five PCFG variants.
We observe the same phenomenon explained in Section~\ref{sec:corpus-level-correlation} for CPCFG, NPCFG, SNPCFG, and TNPCFG. 
The correlation coefficients for (SemInfo, \csfo) are consistently above 0.75, whereas the coefficients for (LL, \csfo) drop quickly as the training progresses.
We can observe the stronger correlation between SemInfo and \csfo in Figure~\ref{fig:training-dynamics}. 
The figure plots the training curves of the corpus-level \csfo score, the average SemInfo value, and the average LL value over the English validation set.
For example, we can see that SemInfo ranks the NPCFG models represented by the green and grey lines as the lowest and those represented by the purple and blue lines as the highest.
This largely agrees with the \csfo scores, where the NPCFG models represented by the green and grey lines are among the bottom three worst-performing models, and the models represented by the blue and purple lines are among the top three best-performing models.
In comparison, we see that all models have similar LL scores, which indicates LL's inability to rank models in accordance with their parsing performance.
These results underscore our conclusion that SemInfo ranks PCFG models better than LL.

In Figure~\ref{fig:corr-corpus-training_stages}, we observe that the correlation strength for (SemInfo, \csfo) also drops as training progresses in SCPCFG.
One reason is that SCPCFG fails to explore constituent structures with high SemInfo values. 
As shown in Figure~\ref{fig:training-dynamics}, the average SemInfo value across the eight models is around 42 for SCPCFG, while the average SemInfo value is greater or equal to 45 for the other four PCFG variants.
This result indicates that the constituent information provided in low SemInfo regions might contain more noise than the information provided in high SemInfo regions.
% This suggests that the SemInfo value might be noisy if 

\subsection{Robustness against Paraphrasing Noises}
\label{appendix:robust-paraphrasing-noise}
\begin{table}[t]
    \centering
    \adjustbox{width=1\textwidth}{
    \begin{tabular}{l|ccc|cc|cc}
    \hline
        ~&\multicolumn{7}{|c}{Paraphrasing Model Variations}\\\cline{2-8}
        ~&\multicolumn{3}{|c|}{Large Models} & \multicolumn{2}{c|}{Medium Models} & \multicolumn{2}{c}{Small Models}\\\cline{2-8}
        ~ & gpt35 & gpt4o & gpt4omini & llama3.2-3b & qwen2.5-3b & llama3.2 1b & qwen2.5-0.5b  \\ \hline
        SemInfo-NPCFG & 66.85±0.25 & 65.19±0.54 & 64.45±1.13 & 63.78±0.55 & 63.58±0.13 & 63.10±0.70 & 59.01±0.24  \\ 
        SemInfo-MTD & 55.56 & 59.45 & 58.28 & 55.17 & 55.03 & 48.5 & 43.3  \\\hline
        LL-NPCFG & \multicolumn{7}{c}{50.96±1.82} \\ 
        Right Branching & \multicolumn{7}{c}{38.4}\\\hline
    \end{tabular}}
    \caption{\csfo\nspace of the NPCFG and MaxTreeDecoding (MTD) parsers using SemInfo values obtained from seven paraphrasing models. LL-NPCFG indicates the \csfo\nspace score of the LL-trained NPCFG parser. }
    \label{tbl:LLM-variation}
\end{table}
Table~\ref{tbl:LLM-variation} compares the parsing accuracy of NPCFG models trained using seven paraphrasing models. 
These models are split into three groups: large models (\texttt{gpt4o}, \texttt{gpt-4o-mini}, \texttt{gpt-3.5}), medium models (\texttt{llama3.2-3b} and \texttt{qwen2.5-3b}), and small models (\texttt{llama3.2-1b} and \texttt{qwen2.5-0.5b}), each representing paraphrasing models with different levels of noises.
The table also includes a MaximumTreeDecoding (MTD) parser, an LL-trained NPCFG parser, and a trivial right-branching parser for reference.
We use the MTD parser to reflect the paraphrasing quality because its parsing accuracy depends solely on the paraphrasing quality.

We can observe that the SemInfo-trained NPCFG parsers are robust against paraphrasing noises.
The accuracy gap between the best (\texttt{gpt4o}) and the worst (\texttt{qwen2.5-0.5b}) performing MTD parser is 16.15 \csfo\nspace score. 
In comparison, the gap between the best and worst performing SemInfo-trained NPCFG parser is 7.84  \csfo\nspace score, less than half of the gap in the MTD parser.
In addition, we can observe that the PCFG parser can benefit from the SemInfo maximization training, even when using noisy paraphrases.
All SemInfo-trained PCFG parsers significantly outperform their LL-trained counterparts by a large margin.
When trained with the most noisy paraphrasing model (\texttt{qwen2.5-0.5b}), the SemInfo-trained PCFG parser outperforms its LL-trained counterpart by 9 points.
The result suggests that the PCFG model effectively suppresses the paraphrasing noise, leading to robust PCFG parsers.

\subsection{Potential for Ensembling}
\label{sec:appendix-ensemble}
\begin{figure}[t]
    \centering
    \begin{subfigure}{0.3\linewidth}
        \centering
        \includegraphics[width=\linewidth]{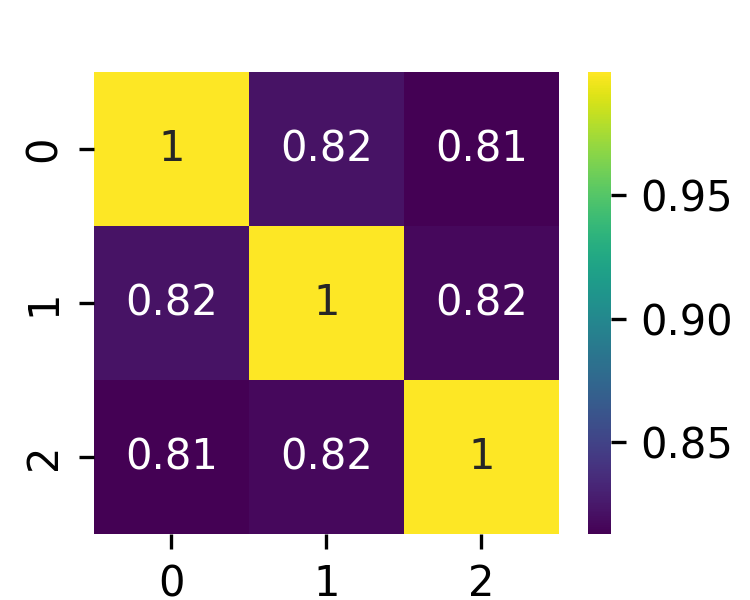}
        \caption{NPCFG}
    \end{subfigure}
    \begin{subfigure}{0.3\linewidth}
        \centering
        \includegraphics[width=\linewidth]{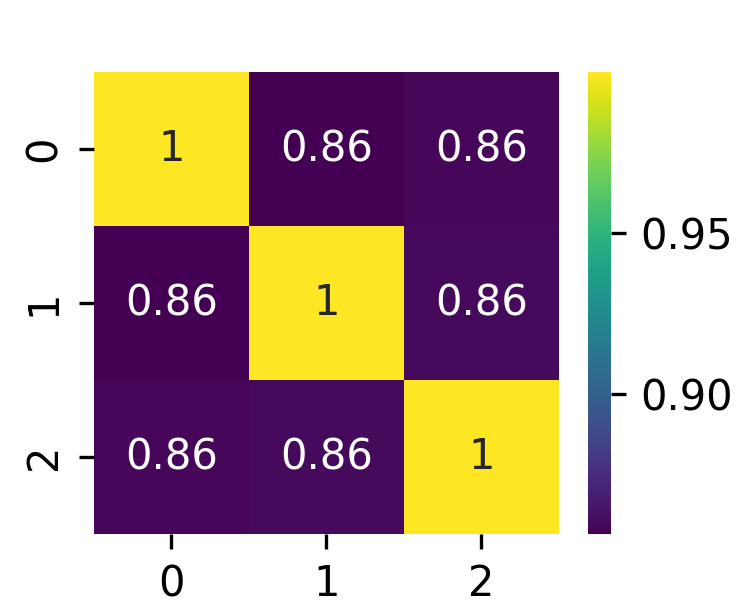}
        \caption{SNPCFG}
    \end{subfigure}
    \begin{subfigure}{0.3\linewidth}
        \centering
        \includegraphics[width=\linewidth]{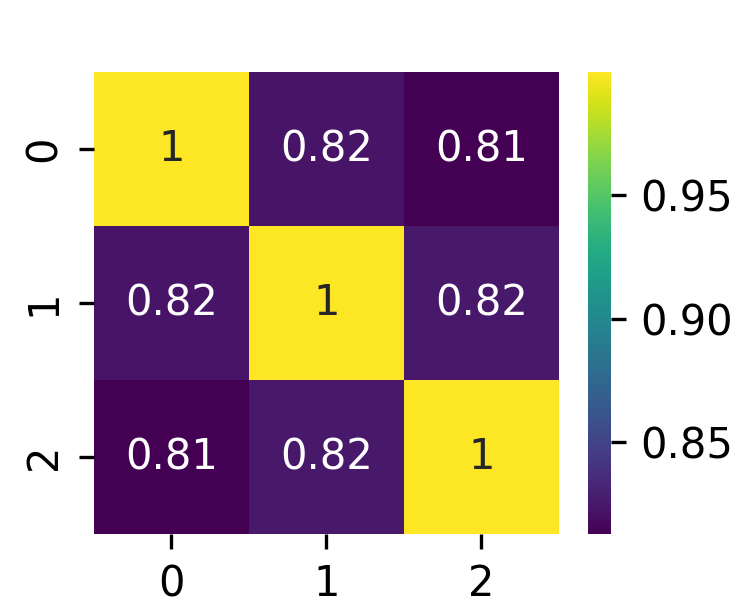}
        \caption{CPCFG}
    \end{subfigure}
    \begin{subfigure}{0.3\linewidth}
        \centering
        \includegraphics[width=\linewidth]{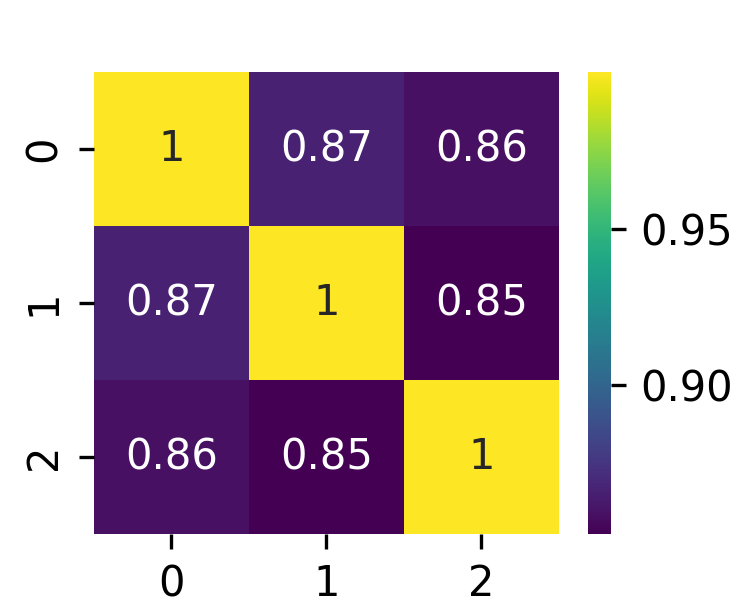}
        \caption{SCPCFG}
    \end{subfigure}
    \begin{subfigure}{0.3\linewidth}
        \centering
        \includegraphics[width=\linewidth]{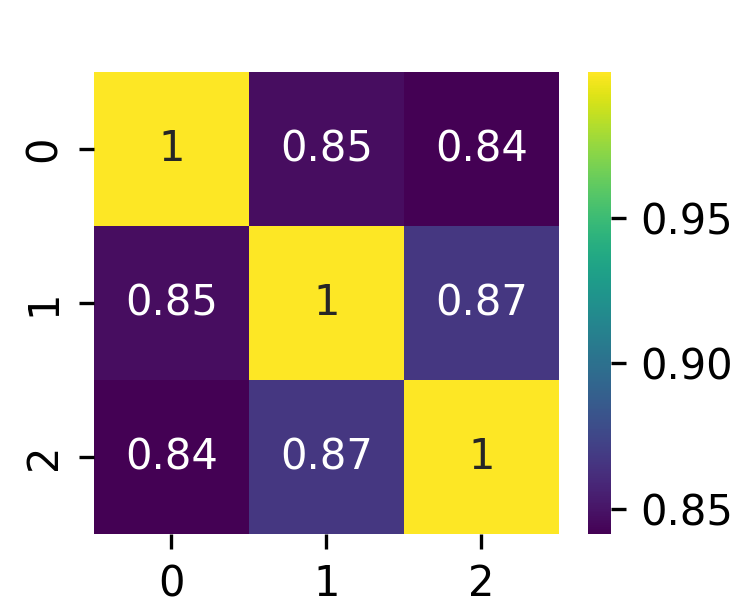}
        \caption{TNPCFG}
    \end{subfigure}
    \caption{PCFG agreements between independent training runs.}
    \label{fig:corr-intraparser}
\end{figure}

\begin{figure}[t]
    \centering
    \includegraphics[width=0.6\linewidth]{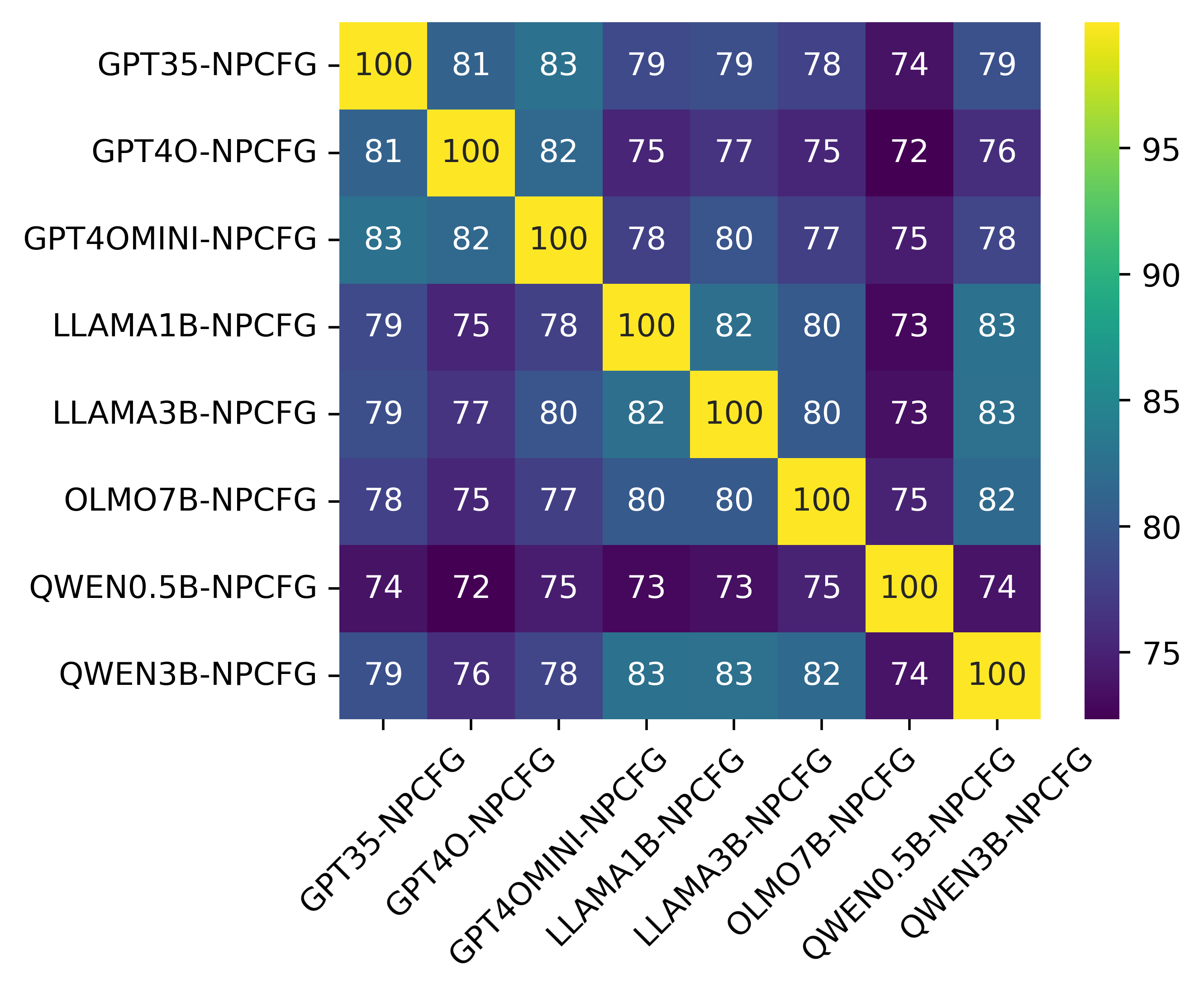}
    \caption{NPCFG parser agreement when trained with different paraphrasing models}
    \label{fig:corr-interLLM}
\end{figure}
\begin{figure}[p]
    \centering
    \includegraphics[width=\linewidth]{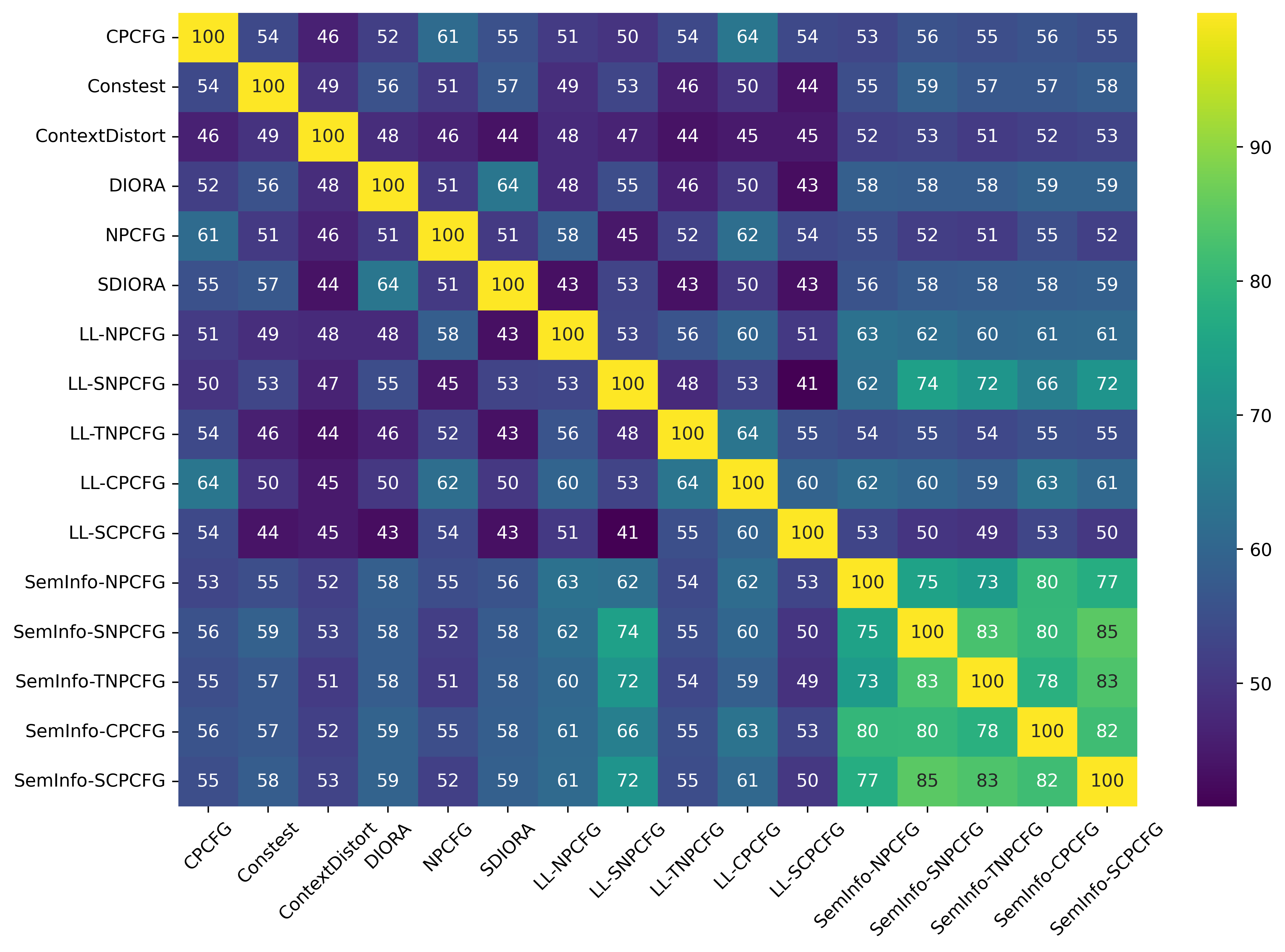}
    \caption{Agreement between heterogeneous parsers}
    \label{fig:corr-interparser}
\end{figure}

Figure~\ref{fig:corr-interLLM}, and Figure~\ref{fig:corr-interparser} suggests that the SemInfo-trained PCFG would benefit from parser ensembling \citep{DBLP:conf/iclr/ShayeghC0CM24}.
\citet{DBLP:conf/iclr/ShayeghC0CM24} shows that homogeneous unsupervised parsers (same parser model, different initializations) make mildly distinctive predictions, and heterogeneous parsers (different parser models) make considerably distinctive predictions. 
Ensembling the parsing results from those parsers effectively suppresses parsing errors made by individual parsers, leading to significant accuracy improvement.

In this section, we evaluate whether our SemInfo-trained PCFGs can benefit from parser ensembling by examining the parser agreement scores for our parser and comparing the score with those reported in \citet{DBLP:conf/iclr/ShayeghC0CM24}.
If our parser exhibits similar agreement scores, we can consider that our parser would benefit from the parser ensembling.
We evaluate the agreement score of our parser and six previous heterogeneous parsers (\texttt{CPCFG}, \texttt{Constest}, \texttt{ContextDistort}, \texttt{DIORA}, \texttt{NPCFG}, and \texttt{SDIORA}).

Figure~\ref{fig:corr-interLLM} illustrates the agreement score among parsers using different paraphrasing models.
The agreement scores (70-83) are similar to the reported score between homogeneous parsers (74-75 \citep{DBLP:conf/iclr/ShayeghC0CM24}).
This similarity in score suggests that the SemInfo-trained PCFG parsers would benefit from ensembling parsers using various paraphrasing models.

Figure~\ref{fig:corr-interparser} illustrates the agreement score among SemInfo-trained PCFG parsers and previous heterogeneous parsers. 
We can observe that the agreement score between our SemInfo-trained PCFG parsers and previous parsers ranges from 54-58 (shown in the top-right corner of Figure~\ref{fig:corr-interparser}).
The score falls in the same range as the score among those previous parsers (46-61, shown in the top-left corner of Figure~\ref{fig:corr-interparser}).
Both scores are similar to the reported heterogeneous agreement score (55-63 \citep{DBLP:conf/iclr/ShayeghC0CM24}). 
This similarity in score suggests that the SemInfo-trained PCFG parsers would benefit from being ensembled with previous heterogeneous parsers.

\subsection{Recall on six most-frequent constituent types}
\begin{table}[t]
    \centering
    \begin{adjustbox}{width=\textwidth}
         \begin{tabular}{l|cc|cc|cc|cc|cc|c}
    \hline
        ~& \multicolumn{2}{|c|}{CPCFG} & \multicolumn{2}{|c|}{NPCFG}  & \multicolumn{2}{|c|}{SCPCFG} & \multicolumn{2}{|c|}{SNPCFG} & \multicolumn{2}{|c|}{TNPCFG} &  \multirow{2}{*}{$\Delta$ by Type} \\ \cline{2-11}
         & SemInfo (Ours) & LL & SemInfo & LL & SemInfo & LL & SemInfo & LL &SemInfo & LL & ~ \\ \hline
        NP & 88.88±\textsubscript{0.06} & 79.77±\textsubscript{1.58} & 88.98±\textsubscript{0.34} & 80.63±\textsubscript{2.10} & 87.45±\textsubscript{1.16} & 79.41±\textsubscript{1.47} & 86.51±\textsubscript{0.18} & 70.95±\textsubscript{1.64} & 87.89±\textsubscript{1.23} & 77.73±\textsubscript{5.72} & +10.90 \\ 
        VP & 71.19±\textsubscript{1.10} & 40.79±\textsubscript{1.49} & 65.69±\textsubscript{2.06} & 28.29±\textsubscript{3.24} & 73.80±\textsubscript{1.65} & 28.53±\textsubscript{1.15} & 76.35±\textsubscript{2.18} & 80.21±\textsubscript{0.51} & 72.23±\textsubscript{2.19} & 45.82±\textsubscript{7.52} & +26.65 \\ 
        PP & 68.22±\textsubscript{5.68} & 72.27±\textsubscript{0.47} & 70.15±\textsubscript{5.42} & 75.15±\textsubscript{0.83} & 79.75±\textsubscript{0.57} & 73.83±\textsubscript{8.94} & 80.26±\textsubscript{1.45} & 78.85±\textsubscript{0.98} & 78.51±\textsubscript{0.83} & 71.07±\textsubscript{8.49} & +2.09 \\ 
        SBAR & 80.99±\textsubscript{1.40} & 52.18±\textsubscript{2.15} & 80.37±\textsubscript{3.48} & 56.32±\textsubscript{6.03} & 84.16±\textsubscript{0.56} & 40.81±\textsubscript{12.99} & 82.17±\textsubscript{0.91} & 81.28±\textsubscript{1.06} & 82.45±\textsubscript{1.55} & 54.46±\textsubscript{4.92} & +22.67 \\ 
        ADVP & 91.87±\textsubscript{0.56} & 88.38±\textsubscript{0.97} & 91.48±\textsubscript{0.61} & 89.78±\textsubscript{1.17} & 92.22±\textsubscript{1.01} & 88.57±\textsubscript{4.53} & 92.11±\textsubscript{0.74} & 89.67±\textsubscript{0.93} & 90.93±\textsubscript{1.59} & 88.07±\textsubscript{0.71} & +4.48 \\ 
        ADJP & 71.82±\textsubscript{1.43} & 63.08±\textsubscript{1.90} & 75.18±\textsubscript{2.85} & 61.66±\textsubscript{9.97} & 78.39±\textsubscript{1.78} & 60.40±\textsubscript{8.03} & 75.77±\textsubscript{3.74} & 75.55±\textsubscript{2.18} & 72.90±\textsubscript{4.19} & 65.40±\textsubscript{6.60} & +7.93 \\ \hline
        $\Delta$ by Model & \multicolumn{2}{|c|}{+12.42}  & \multicolumn{2}{|c|}{+13.05}  & \multicolumn{2}{|c|}{+20.14}  & \multicolumn{2}{|c|}{+3.90}  & \multicolumn{2}{|c|}{+12.76}  & ~ \\ \hline
    \end{tabular}   
    \end{adjustbox}
    \caption{Recall on six most frequent constituent types. The recall data is calculated over the English test set. $\Delta$ by Type indicates the average recall improvement for the constituent type. $\Delta$ by Model indicates the average recall improvement for the PCFG variant.}
    \label{tbl:recall}
   
\end{table}
Table~\ref{tbl:recall} shows the recall of the six most frequent constituent types on the English test set, following \citet{yang-etal-2021-pcfgs}.
We see that PCFGs trained with SemInfo achieve significant improvement in Noun Phrases (NP), Verb Phrases (VP), and Subordinate Clauses (SBAR). 
These three constituents are the most typical constituents that carry semantic information. 
The significant improvement underscores the importance of semantic information in identifying the constituent structure.

\subsection{Paraphrasing prompts}
\label{appendix:prompts}
We use the below prompts to generate paraphrases from the \texttt{gpt-4o-mini-2024-07-18} model. \{lang\} is a placeholder for languages. For example, we set \{lang\}=``English'' when collecting English paraphrases.
\begin{itemize}
    \item Create grammatical sentences by shuffling the phrases in the below sentence. The generated sentences must be in \{lang\}. Use the same word as in the original sentence
\item Create grammatical sentences by changing the tense in the below sentence. The generated sentences must be in \{lang\}. Use the same word as in the original sentence.
\item Create grammatical sentences by restating the below sentences in passive voice. The generated sentences must be in \{lang\}. Use the same word as in the original sentence. 
\item Create grammatical sentences by restating the below sentences in active voice. The generated sentences must be in \{lang\}. Use the same word as in the original sentence. 
\item Create grammatical clefting sentences based on the below sentence. The generated sentences must be in \{lang\}. Use the same word as in the original sentence.
\item Create pairs of interrogative and its answers based on the below sentence. The generated sentences must be grammatically correct and be explicit. The sentences must be in \{lang\}. Use the same word as in the original sentence. The answer to the questions should be a substring of the given sentence.
\item Create pairs of confirmatory questions and its answers based on the below sentence. The generated sentences must be grammatically correct and textually diverse. The sentences must be in \{lang\}. Use the same word as in the original sentence. The answer to the questions should be a substring of the given sentence.
\item Create grammatical sentences by performing the topicalization transformation to the below sentence. The sentences must be in \{lang\}. Use the same word as in the original sentence. 
\item Create grammatical sentences by performing the heavy NP shift transformation to the below sentence. The sentences must be in \{lang\}. Use the same word as in the original sentence.
\end{itemize}

\subsection{Examples of the collected paraphrases}
The below list contains examples of our collected paraphrases for \textit{Such agency ` self-help ' borrowing is unauthorized and expensive , far more expensive than direct Treasury borrowing , said Rep. Fortney Stark -LRB- D. , Calif. -RRB- , the bill 's chief sponsor .}.
\begin{itemize}
\item 'Self-help' borrowing by such agency is unauthorized and expensive, far more expensive than direct Treasury borrowing,' said Rep. Fortney Stark -LRB- D., Calif. -RRB-, the bill's chief sponsor.
\item Far more expensive than direct Treasury borrowing is such agency ' self-help ' borrowing, unauthorized and expensive, said Rep. Fortney Stark -LRB- D., Calif. -RRB-, the bill 's chief sponsor.
\item Yes, he said it is far more expensive than direct Treasury borrowing.
\item What is unauthorized and expensive is such agency 'self-help' borrowing, far more expensive than direct Treasury borrowing, according to Rep. Fortney Stark.
\item 'Self-help' borrowing by such agency is considered unauthorized and is regarded as expensive, far more expensive than direct Treasury borrowing,'' said Rep. Fortney Stark -LRB- D., Calif. -RRB-, who is the chief sponsor of the bill.
\item According to Rep. Fortney Stark -LRB- D. , Calif. -RRB- , the bill 's chief sponsor , such agency 'self-help' borrowing is unauthorized and far more expensive than direct Treasury borrowing.
\end{itemize}

\end{document}